\documentclass{article}

\usepackage[final,nonatbib]{neurips_2024}

\usepackage[utf8]{inputenc} % allow utf-8 input
\usepackage[T1]{fontenc}    % use 8-bit T1 fonts
\usepackage{url}            % simple URL typesetting
\usepackage{booktabs}       % professional-quality tables
\usepackage{amsfonts}       % blackboard math symbols
\usepackage{nicefrac}       % compact symbols for 1/2, etc.
\usepackage{microtype}      % microtypography
\usepackage{xcolor}         % colors
\usepackage{cite}
\usepackage{makecell}
\usepackage{multirow}
\usepackage{amsmath}
\usepackage{tikz}
\usepackage{enumitem}
\usepackage{subcaption}
\usepackage{graphicx}  
\definecolor{eccvblue}{rgb}{0.12,0.49,0.85}
\usepackage[pagebackref,breaklinks,colorlinks,citecolor=eccvblue]{hyperref}
\definecolor{fg}{rgb}{0.88, 0.11, 0.09} 
\definecolor{bg}{rgb}{0.49, 0.8, 0.51} 
\definecolor{myred}{RGB}{162, 50, 35}
\definecolor{mygreen}{RGB}{134, 183, 55}
\definecolor{myblue}{RGB}{34, 34, 199}
\definecolor{myorange}{RGB}{223, 134, 82}

\title{RadarOcc: Robust 3D Occupancy Prediction with \\ 4D Imaging Radar}

\author{%
  \textbf{Fangqiang Ding}\textsuperscript{1,}\thanks{Equal contribution}\quad
  \textbf{Xiangyu Wen}\textsuperscript{1,}\footnotemark[1] \quad
  \textbf{Yunzhou Zhu}\textsuperscript{2} \quad
  \textbf{Yiming Li}\textsuperscript{3} \quad
  \textbf{Chris Xiaoxuan Lu}\textsuperscript{4}\thanks{Corresponding author. Email: xiaoxuan.lu@ucl.ac.uk} \\
  \textsuperscript{1}University of Edinburgh \quad
  \textsuperscript{2}Georgia Institute of Technology \quad
  \textsuperscript{3}New York University \quad\\
  \textsuperscript{4}AI Centre, Department of Computer Science, UCL \\
}

\newcommand{\sysname}{\texttt{RadarOcc}}

\begin{document}

\maketitle

\begin{abstract}
3D occupancy-based perception pipeline has significantly advanced autonomous driving by capturing detailed scene descriptions and demonstrating strong generalizability across various object categories and shapes. Current methods predominantly rely on LiDAR or camera inputs for 3D occupancy prediction. These methods are susceptible to adverse weather conditions, limiting the all-weather deployment of self-driving cars. To improve perception robustness, we leverage the recent advances in automotive radars and introduce a novel approach that utilizes 4D imaging radar sensors for 3D occupancy prediction. Our method, RadarOcc, circumvents the limitations of sparse radar point clouds by directly processing the 4D radar tensor, thus preserving essential scene details. RadarOcc innovatively addresses the challenges associated with the voluminous and noisy 4D radar data by employing Doppler bins descriptors, sidelobe-aware spatial sparsification, and range-wise self-attention mechanisms. To minimize the interpolation errors associated with direct coordinate transformations, we also devise a spherical-based feature encoding followed by spherical-to-Cartesian feature aggregation. We benchmark various baseline methods based on distinct modalities on the public K-Radar dataset. The results demonstrate RadarOcc's state-of-the-art performance in radar-based 3D occupancy prediction and promising results even when compared with LiDAR- or camera-based methods. Additionally, we present qualitative evidence of the superior performance of 4D radar in adverse weather conditions and explore the impact of key pipeline components through ablation studies.
\end{abstract}

\section{Introduction}
The safety of autonomous vehicles navigating in the wild hinges on a thorough understanding of the environment's 3D structure. As a unified scene representation built from grid-based volumetric elements known as voxels, 3D occupancy has gained increasing attention within the autonomous driving community~\cite{youtube2022tesla,agro2023implicit,wang2023openoccupancy,tian2024occ3d,li2023voxformer}. Its rising popularity stems from its comprehensive scene depiction, capturing both geometric and semantic aspects. Crucially, it transcends the limitations of foreground-only representations (\emph{vs.} 3D object detection~\cite{liang2022bevfusion,liu2023bevfusion,chen2023futr3d}) and sparse data formats (\emph{vs.} point cloud segmentation~\cite{kong2023rethinking,chen2021polarstream,lai2022stratified}). Furthermore, 3D occupancy offers a detailed open-set depiction of scene geometry, effectively handling out-of-vocabulary items (e.g., animals) and irregular shapes (e.g., cranes). This capability allows it to address a broader range of corner cases than previous object-based perception approaches~\cite{liang2020pnpnet,gu2023vip3d,hu2023planning}.

Previous research has predominantly utilized either LiDAR point clouds~\cite{cheng2021s3cnet,roldao2020lmscnet,yan2021sparse,li2021semi,rist2021semantic,agro2023implicit,liu2023lidar,xia2023scpnet,khurana2023point}, RGB images~\cite{cao2022monoscene,huang2023tri,li2023voxformer,wei2023surroundocc,tong2023scene,zhang2023occformer,tian2024occ3d,tan2023ovo,huang2023selfocc,zhang2023occnerf,vobecky2024pop,ma2023cotr,ma2023cam4docc}, or a combination of both~\cite{wang2023openoccupancy} for 3D occupancy prediction. However, the potential of 4D imaging radar~\cite{sun20214d,sun2020mimo} —a critical sensor in autonomous driving—has been largely untapped in this area. Evolving from traditional 3D mmWave radars, this emerging sensor type enhances elevation resolution, enabling detection and resolution of targets across both horizontal and vertical planes, which results in detailed \emph{imaging} outputs. Meanwhile, 4D radar inherits the traditional advantages of mmWave radar, such as capability in all lighting and weather conditions, object velocity measurement, and cost-effectiveness compared to LiDAR systems. These attributes, particularly its resilience in adverse weather conditions like fog and rain, position 4D radar as an essential component in achieving mobile autonomy.

% \begin{figure}[!tbp]
%     \centering
%     % \includegraphics[width=\textwidth]{path/to/your/image.png}
%     \fbox{\rule[-.5cm]{0cm}{2cm} \rule[-.5cm]{2cm}{0cm}}
%     \caption{\text{We need an open figure here.}}
%     \label{fig:openfigure}
%     \vspace{-2em}
% \end{figure}

In this work, we explore the potential of 4D imaging radar to enhance 3D occupancy prediction. Previous research in radar perception has largely relied on 4D radar point clouds as input, a method inspired by LiDAR techniques. This `LiDAR-inspired' framework has demonstrated effectiveness in tasks such as 3D object detection and tracking~\cite{meyer2019automotive,liu2024echoes,Rebut_2022_CVPR,xu2021rpfa,palffy2022multi,tan20223d,paek2022k,meyer2019deep,wang2022interfusion,wang2022multi,paek2023enhanced,zheng2022tj4dradset,liu2023smurf,zheng2023rcfusion,xiong2023lxl,zhang2024tl,kong2023rtnh+,yan2023mvfan,zhang2023dual,deng2023see,cui20213d,pan2023moving,tan2023tracking}. However, this approach primarily enhances the detection of foreground objects such as cars, pedestrians, and trucks. In contrast, 3D occupancy prediction requires the detection of signal reflections from all occupied spaces, encompassing both foreground and background elements like roads, barriers, and buildings. The traditional reliance on sparse radar point clouds, therefore, is not optimal for 3D occupancy prediction, as critical environmental signals are often lost during the point cloud generation process \cite{scharf1991statistical,gandhi1988analysis}. For instance, the surface of highways, typically made of low-reflectivity materials such as asphalt, often yields weak signals back to the radar receiver.

To avoid the loss of negligible signal returns, we propose utilizing the 4D radar tensor (4DRT) for 3D occupancy prediction. This raw data format preserves the entirety of radar measurements, offering a comprehensive dataset for analysis. However, employing such volumetric data introduces significant challenges. For instance, the substantial size of 4DRTs—potentially up to 500MB—poses processing inefficiencies that could compromise real-time neural network performance. Additionally, raw radar data is inherently noisy due to the multi-path effect and is stored in spherical coordinates, which diverges from the preferred 3D Cartesian occupancy grid used in our applications.

Motivated by the outlined challenges, we introduce a novel approach, {\sysname}, specifically tailored for 4DRT-based 3D occupancy prediction. To address the computational and memory demands, our method initially reduces the data volume of 4DRTs through the encoding of Doppler bins descriptors and implementing spatial sparsification in the preprocessing stages. Our technique features sidelobe-aware spatial sparsification to minimize the interference scattered across azimuth and elevation axes, which is further refined through range-wise self-attention mechanisms. Importantly, we observed the typical conversion of spherical RTs to Cartesian data volumes, which often incurs non-negligible interpolation errors. Instead, we directly encode spatial features in spherical coordinates and seamlessly aggregate them using learnable voxel queries defined in Cartesian coordinates. Our approach further employs 3D sparse convolutions and deformable attention~\cite{zhu2020deformable} for efficient feature encoding and aggregation. {\sysname} is benchmarked on the K-Radar dataset~\cite{paek2022k} against state-of-the-art methods across various modalities, demonstrating the promising performance in radar-based 3D occupancy prediction. Comprehensive experiment results validate its comparable performance to the camera and LiDAR solutions. A qualitative assessment further validates the superior robustness of 4D radar data under adverse weather conditions, establishing its capability for all-weather 3D occupancy prediction. The contributions of this work are three-fold:

\vspace{-0.5em}
\begin{itemize}
\setlength{\itemsep}{0pt}
\setlength{\parsep}{0pt}
\setlength{\parskip}{0pt}
    \item Introduction of the first-of-it-kind method, {\sysname}, for 4D radar-based 3D occupancy prediction in autonomous driving. We recognize the limitation of radar point clouds in reserving critical raw signals and advocate the usage of 4DRT for occupancy perception. 
    \item Development of a novel pipeline with techniques to cope with challenges accompanying 4DRTs, including reducing large data volume, mitigating sidelobes measurements and interpolation-free feature encoding and aggregation.
    \item Extensive experiments on the K-Radar dataset, benchmarking state-of-the-art methods based on different modalities, and validating the competitive performance of {\sysname} and its robustness against adverse weather. We release our code and model at \url{https://github.com/Toytiny/RadarOcc}.
\end{itemize}
\section{Related work}
\noindent\textbf{3D occupancy prediction.} Early attempts on 3D occupancy prediction, \emph{aka.} semantic scene completion (SSC)~\cite{roldao20223d}, are mainly limited to the small-scale interior scenes~\cite{song2017semantic, roldao20223d,liu2018see,zhang2018efficient,li2019depth,li2019rgbd,zhang2019cascaded,li2020anisotropic,chen20203d,cai2021semantic}. The introduction of SemanticKITTI~\cite{behley2019semantickitti} expands the study of SSC to large-scale outdoor scenes, based on which some works validate the feasibility of outdoor SSC with LiDAR input~\cite{cheng2021s3cnet,roldao2020lmscnet,yan2021sparse,li2021semi,rist2021semantic}. In contrast, MonoScene~\cite{cao2022monoscene} is the seminal work for SCC using only a single monocular RGB image.  
Since Tesla's disclosure of their occupancy network for Full Self-Driving (FSD)~\cite{youtube2022tesla}, there has been a recent surge of research on 3D occupancy prediction for autonomous vehicles. While a few works leverage LiDAR point clouds~\cite{wang2023openoccupancy,agro2023implicit,liu2023lidar,xia2023scpnet,khurana2023point} for scene completion, the majority of existing approaches rely on a vision-only pipeline that learns to lift 2D features into the 3D space~\cite{wang2023openoccupancy,huang2023tri,li2023voxformer,wei2023surroundocc,tong2023scene, zhang2023occformer,tian2024occ3d,tan2023ovo,huang2023selfocc,zhang2023occnerf,vobecky2024pop,ma2023cotr,ma2023cam4docc}. Despite these prevalent solutions based on LiDAR and camera, 4D radar sensors are still under-explored for 3D occupancy prediction. 
% \vspace{-1em}

\noindent\textbf{4D radar for autonomous driving.} As an emerging automotive sensor, 4D mmWave radar prevails over LiDAR and camera in adverse weather (\emph{e.g.}, fog, rain and snow), offering all-weather sensing capabilities for mobile autonomy. In recent years, increasing endeavours have been witnessed to unveil the potential of 4D radar for autonomous driving applications, encompassing 3D object detection~\cite{meyer2019automotive,liu2024echoes,Rebut_2022_CVPR,xu2021rpfa,palffy2022multi,tan20223d,paek2022k,meyer2019deep,wang2022interfusion,wang2022multi,paek2023enhanced,zheng2022tj4dradset,liu2023smurf,zheng2023rcfusion,xiong2023lxl,zhang2024tl,kong2023rtnh+,yan2023mvfan,zhang2023dual,deng2023see,cui20213d,pan2023moving} and tracking~\cite{cui20213d,pan2023moving,tan2023tracking}, scene flow estimation~\cite{ding2023hidden,ding2022self,Ding_2024_ECCV}, odometry~\cite{zhang2023ntu4dradlm,ding2023hidden,choi2023msc,lu2023efficient,zhuoins20234drvo,zhuang20234d,li20234d,zhang20234dradarslam} and mapping~\cite{zhuang20234d,li20234d,zhang20234dradarslam}. Apart from these works, we are the pioneering study for 4D radar-based 3D occupancy prediction, further exploring this unique sensor for the untouched topic.
\vspace{-1em}
\paragraph{Radar tensor for perception} Besides the post-processing radar point cloud, another data type of mmWave radar is the radar tensor (RT), which is the product of applying FFT along the corresponding dimensions to the raw ADC samples (\emph{c.f.} Sec.~\ref{signal}). Unlike the sparse radar point cloud, dense RTs contain rich and complete measurements of the environment, refraining from information loss during point cloud generation (\emph{e.g.}, CFAR~\cite{scharf1991statistical,gandhi1988analysis}). Consequently, some works attempt to use 2D~\cite{rebut2022raw,zhang2020object,wang2021rodnet,dong2020probabilistic,liu2024echoes}, 
3D~\cite{zhang2021raddet,palffy2020cnn,major2019vehicle} or 4D~\cite{paek2023enhanced,paek2022k,kong2023rtnh+} RTs for object detection, yielding satisfactory performance. In this work, we develop a tailored approach to 4D radar-based 3D occupancy prediction based on 4DRTs.
\section{Preliminary}\label{preliminary}
% In this work, our approach is based on the commercial 4D frequency-modulated continuous-wave (FMCW) radars that can provide the measurements along four dimensions (\emph{i.e.,} range, azimuth, elevation and Doppler). 
% To aid in comprehending our method, this section first provides an overview of the basic signal processing pipeline of 4D FMCW radar (\emph{c.f.} Sec.~\ref{signal}), explaining the sources of different data representations of mmWave radar. Following this, we analyze the rationale of using 4DRTs for 3D occupancy prediction and discuss the associated challenges  (\emph{c.f.} Sec.~\ref{4DRT}). 

\subsection{4D radar signal processing pipeline}\label{signal}

\noindent\textbf{ADC samples.} To measure the surroundings, a sequence of FMCW waveforms, aka. chirp signals, are emitted by the transmit (Tx) antennas within a short timeframe. These signals are reflected off objects and captured by the receive (Rx) antennas. The intermediate frequency (IF) signal is produced by mixing the signals from a Tx-Rx antenna pair. This mixed signal is then sampled by an Analog-to-Digital Converter (ADC) to generate discrete samples for each chirp~\cite{TIMmWaveRadar2024}. By compiling ADC samples from all chirps and Tx-Rx antenna pairs, the FMCW radar system constructs a 3D complex data cube for each frame. This data cube is organized into three dimensions: \emph{fast time}, \emph{slow time}, and \emph{channel}, which correspond to range, range rate, and angle, respectively~\cite{kramer2022coloradar}.
% Here, $N_s$ is the number of samples per chirp, $N_c$ is the number of transmitted chirps within one frame and $N_p$ is the number of Tx-Rx pairs.

\noindent\textbf{Radar tensor.} Utilizing ADC samples, Fast Fourier Transforms (FFTs) are applied across relevant dimensions to extract detailed information. The first FFT, known as range-FFT, is performed across the sample (fast time) axis to separate objects at different distances into distinct frequency responses within range bins defined by hardware specifications. Subsequently, a Doppler-FFT along the chirp (slow time) axis decodes phase variances—Doppler bins—to derive relative radial velocities, producing a range-Doppler heatmap. For configurations with multiple Rx-Tx antenna pairs, termed \emph{virtual} antenna elements, additional FFTs (angle-FFT) are executed across the spatial dimensions of the virtual antenna array to determine Angles of Arrival (AoA) for azimuth and elevation angles. This series of transformations results in a comprehensive 4D radar tensor (4DRT), characterized by power measurements across range, Doppler velocity, azimuth, and elevation dimensions. 

%If the antenna array spans only horizontally, a single additional FFT yields a range-Doppler-azimuth mapping. \chris{FQ, do we need the above sentence about `...spans only horizontally...'?}

\noindent\textbf{Radar point cloud.} Beyond analyzing radar tensors, most FMCW radar sensors further refine their output to identify salient targets, which typically represent less than 1\% of the data. Target detection algorithms such as CA-CFAR~\cite{scharf1991statistical} and OS-CFAR~\cite{blake1988cfar} are commonly applied to the range-Doppler heatmap~\cite{cheng2022novel,kramer2022coloradar} or directly on the 3D/4D radar tensors~\cite{paek2022k,paek2023enhanced} to isolate peak measurements. This process generates a sparse radar point cloud, with each point characterized by 3D coordinates and attributes such as Doppler velocity, power intensity, or radar cross-section (RCS). While this step significantly reduces data volume and mitigates noise, it also eliminates a substantial amount of potentially valuable information. 

\subsection{4DRT for 3D occupancy prediction}\label{4DRT}

\noindent\textbf{Rationale of using 4DRT.} 4D radar tensors (4DRTs) serve as raw sensor data that amalgamate the strengths of LiDAR/radar point clouds and RGB images, providing direct 3D measurements in a continuous data format. These tensors comprehensively capture information from raw radar measurements, effectively addressing the shortcomings associated with the sparseness of radar point clouds caused by the signal post-processing. For instance, low-reflectivity surfaces like asphalt, common on highways, typically do not reflect enough radar signals for detection. By using 4DRTs, these minimal signal returns can be detected, significantly bolstering occupancy prediction capabilities. Furthermore, the volumetric structure of 4DRTs aligns well with 3D occupancy grids, making them ideally suited for advancing 3D occupancy prediction techniques.

\noindent\textbf{Challenges.} Despite their significant advantages, using 4D radar tensors (4DRTs) for 3D occupancy prediction presents substantial challenges. First, the large data size of 4DRTs (e.g., 500MB per frame in the K-Radar dataset~\cite{paek2022k}) hinders computational efficiency, necessitating data volume reduction before processing. Second, the inherent noise in radar data, exacerbated by the multi-path effect of mmWave, requires careful filtering to preserve essential signals while eliminating noise. Third, the discrepancy between the spherical coordinates of 4DRT data and the Cartesian coordinates required for 3D occupancy outputs calls for a tailored network design. This design must effectively translate spatial interactions from spherical to Cartesian dimensions to ensure accurate occupancy predictions.
\section{Method}
\subsection{Task definition}
In this work, we consider the task of 3D occupancy prediction with single-frame 4DRT output from 4D imaging radar. Given a 4DRT captured in the current frame denoted as $\mathbf{V}\in\mathbb{R}^{R\times A\times E\times D}$, our task aims to predict a 3D volume $\mathbf{O}=\{o_i\}_{i=1}^{H\times W\times L}$, of which each voxel element $o_i\in\{c_0, c_1, \dots, c_C\}$ is represented as either free (\emph{i.e.,} $c_0$) or occupied with a certain semantics $c_i (i>0)$ out of $C$ classes. Here, $R$, $A$, $E$, and $D$ denote the number of bins along the range, azimuth, elevation and Doppler axis, respectively, and each scalar of the 4DRT is the power measurement mapped to a location within the space defined by these four axes. $H$, $W$ and $L$ represent the volumetric size of the predefined region of interest (RoI) in the height, width and length dimensions. 

\subsection{Overview}
{\sysname} consists of four components in tandem (\emph{c.f.} Fig.~\ref{fig:pipeline}). Before loading heavy 4DRTs to the neural network, we reduce their data volume as the preprocessing steps via encoding the Doppler bins descriptor and performing sidelobe-aware spatial sparsifying to improve the efficiency without losing the key information (\emph{c.f.} Sec.~\ref{reduction}). To refrain from the interpolation error, we encode spatial features directly on the spherical RTs without transforming them into Cartesian volumes (\emph{c.f.} Sec.~\ref{spherical}) and aggregate the spherical features with 3D volume queries defined in the Cartesian coordinates (\emph{c.f.} Sec.~\ref{aggregation}). Specifically, range-wise self-attention is used to alleviate the sidelobes, and sparse convolution and deformable attention are leveraged for fast feature encoding and aggregation. The occupancy probabilities are predicted in the 3D occupancy decoding step, which is supervised via our training loss~(\emph{c.f.} Sec.~\ref{decoding}).

\subsection{Data volume reduction}\label{reduction}
\begin{figure}[!tbp]
    \centering
    \includegraphics[width=\textwidth]{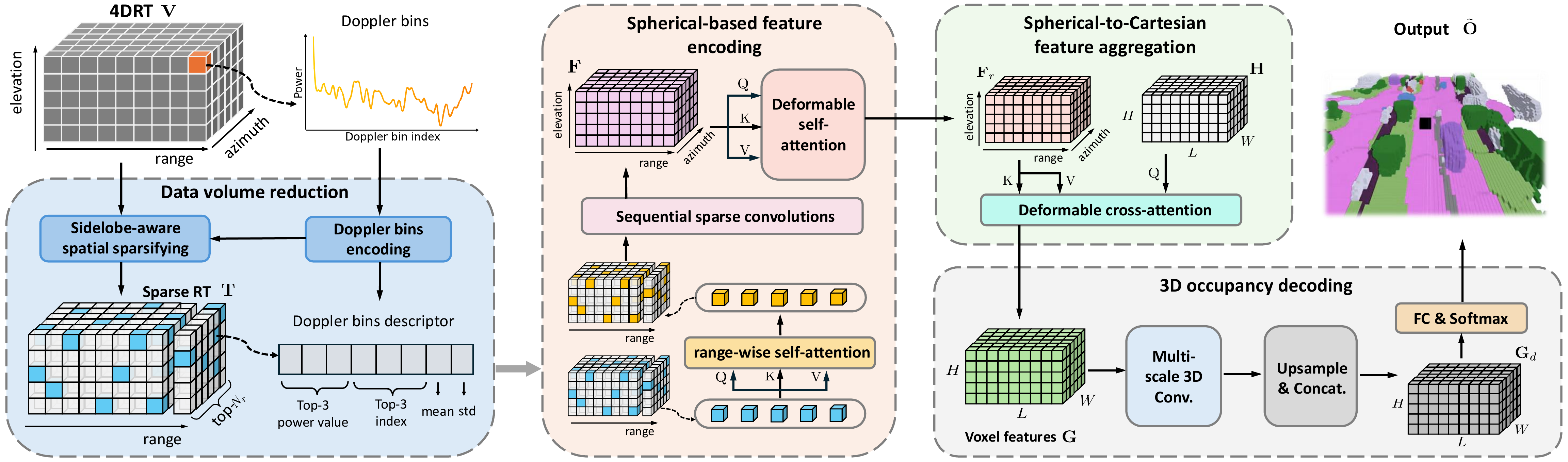}
    % \fbox{\rule[-.5cm]{0cm}{4.5cm} \rule[-.5cm]{4.5cm}{0cm}}
    \caption{Overall pipeline of \sysname. The data volume reduction pre-processes the 4DRT into a lightweight sparse RT via Doppler bins encoding and sidelobe-aware spatial sparifying. We apply spherical-based feature encoding on the sparse RT and aggregate the spherical features using Cartesian voxel queries. The 3D occupancy volume is finally output via 3D occupancy decoding.}
    \label{fig:pipeline}
    \vspace{-1em}
\end{figure}

Direct processing of raw 4DRTs with neural networks is impractical due to its substantial data size (\emph{e.g.}, 500MB per frame) which leads to heavy computation cost and memory usage. Moreover, the slow data transfer between the sensor, storage device and processing unit (CPU/GPU) of large-volume raw 4DRTs not only hinders the onboard runtime efficiency but also increases the training duration which demands repetitive data loading. For efficiency, we propose to reduce the data volume of 4DRTs through encoding the {Doppler bins descriptor} and {sidelobe-aware spatial sparsifying} as the preprocessing steps (see Fig.~\ref{fig:pipeline}). Post reduction, the loading of 4DRTs into the processing unit for runtime inference can be more feasible and the network training can be more efficient. 

\noindent\textbf{Doppler bins descriptor.}
Unlike the three spatial axes, which are intuitively critical for spatial perception, the Doppler axis in 4DRTs has often been considered redundant in 3D object detection. Previous studies~\cite{paek2022k,paek2023enhanced,kong2023rtnh+} have employed average-pooling to minimize this axis, aiming to reduce computational overhead. However, we argue that this ostensibly `redundant' axis contains vital cues for geometric and semantic analysis in 3D occupancy prediction. Specifically, the Doppler axis provides essential information on object speed via peak locations, aiding in differentiating dynamic objects from static backgrounds. Moreover, the power distribution within the Doppler bins offers insights into the confidence levels of true targets—essentially, indicating their likelihood of occupancy. To preserve and utilize this crucial information, we have developed a method to encode the Doppler bins into a descriptor that captures specific statistics for each spatial location within the 4DRTs. This descriptor incorporates the top-three power values along with their indices, the mean power value, and the standard deviation, as depicted in Fig~\ref{fig:pipeline}. Note that the number of preserved top values is determined empirically.
Consequently, this approach enables us to reduce the data volume of raw 4DRTs by a factor of $\frac{D}{8}$, while retaining key information from the Doppler axis.

\begin{figure}[!tbp]
    \centering
    \includegraphics[width=\textwidth]{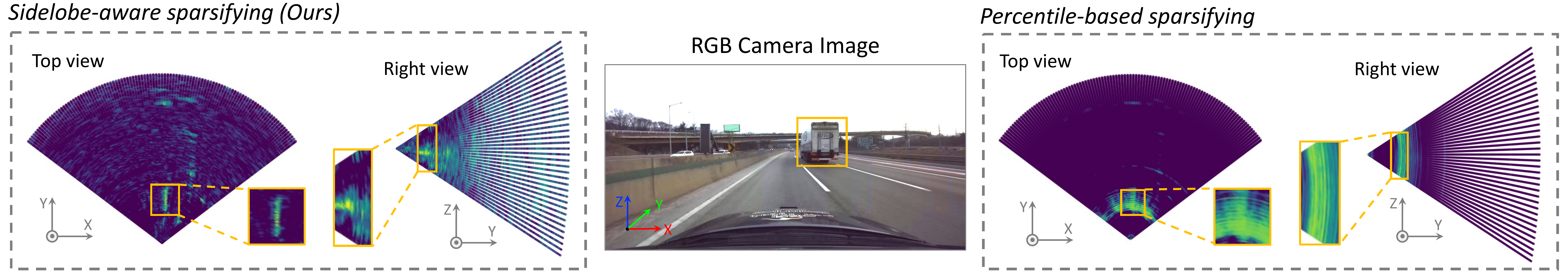}
    \caption{Comparison between the sparse RTs resulted by our sidelobe-aware and percentile-based sparsifying~\cite{paek2023enhanced,paek2022k}. We transform the spherical RT elements to the Cartesian coordinates and show them in two views. The arches on the heatmap indicate the same ranges. Percentile-based method retains many elements caused by sidelobe noise, which are concentrated at certain ranges.  In contrast, our method can reduce the sidelobe level and reserve critical measurement from different ranges. }
    % \chris{the caption is not informative. We need to spell out the exact difference / consequence of using different sparsifying method. For now, I can see the different pictures of the above, but cannot link it to our problem on 3d occupancy prediction.} } 
    \label{fig:sparsifying}
    \vspace{-1em}
\end{figure}

\noindent\textbf{Sidelobe-aware spatial sparsifying.}
By encoding the Doppler bins into light-weight descriptors, we transform the raw 4DRT into 3D spatial data volume 
% $\textbf{M}\in\mathbb{R}^{R\times A\times E\times 8}$
with the original Doppler axis as the 8-channel feature dimension. Nevertheless, it remains costly for neural networks to encode features from 3D dense data volume with operations like 3D convolution~\cite{tran2015learning,zhou2018voxelnet}. To accelerate the computation, prior arts~\cite{paek2023enhanced,paek2022k} transfer the dense RT into a sparse format by retraining only the top-percentile elements based on power measurements. However, this approach tends to be biased towards specific ranges that exhibit exceptionally high measurements. It can be observed in Fig.~\ref{fig:sparsifying} that after percentile-based sparsifying, a significant number of the reserved elements are concentrated within the same ranges spread across the azimuth and elevation axes. These elements manifest as artifacts of sidelobes, which can an be viewed as the diffraction pattern of the antenna~\cite{kwok2015effects,tait2005introduction}. 
% \chris{FQ, the ML reviewers may not know this term. I suggest explaining this sidelobe with 1 or 2 sentences and also cite a radar paper here.} 
Consequently, this results in the loss of important measurements from other ranges and introduces lots of noise into the sparse tensor. To mitigate this issue, we propose to select the top-$N_r$ elements for each individual range instead of on the whole dense RT for spatial sparsifying (see Fig.~\ref{fig:pipeline}). In this way, the dominance of certain ranges can be avoided while the sidelobe level is reduced, as exhibited in Fig.~\ref{fig:sparsifying}. Note that our spatial element selection is based on the mean power value across the Doppler axis. The final sparse tensor is denoted as $\mathbf{T}=\{t_i\in\mathbb{R}^{N_r\times (8+2)}\}_{i=1}^{R}$ with the extra two feature channels storing the azimuth and elevation indices of reserved $N_r$ elements for each range.

% Besides, we change the tensor data from the original 64-bit to 32-bit, which halves the data size but hardly affects the prediction accuracy \chris{this implementation detail can be removed if we need space. }. As a result, we reduce the data volume of raw 4DRTs by a factor of
% $\sim$100 into a lightweight sparse RT, 

\subsection{Spherical-based feature encoding}\label{spherical}
Given the sparse RT, we aim to encode representative features for accurate 3D occupancy prediction. As the sparse RTs are inherently in the spherical coordinates, previous works~\cite{paek2022k,paek2023enhanced} transfer them into the Cartesian coordinates before feature encoding. However, such a transfer would undermine their uniform density distribution and often incur interpolation errors. Inspired by the polar representation of point clouds~\cite{nie2023partner,chen2021polarstream,zhu2021cylindrical}, we propose to take the elements in RT as voxels rasterized in the spherical coordinates and apply the spherical-based feature encoding directly. The spherical voxel representation naturally matches the spherical-uniform distribution of RTs and can refrain from inducing interpolation errors. In practice, the 3D convolutions can be used to extract grid-based representations by only replacing the $X$-$Y$-$Z$ axis with the range-azimuth-elevation axis. In what follows, we illustrate our spherical-based feature encoding process. 

\noindent\textbf{Range-wise self-attention.} In Section~\ref{reduction}, we address the issue of sidelobes by selecting elements based on range-wise percentiles during the preprocessing phase. To further mitigate sidelobe interference, we introduce a range-wise self-attention mechanism~\cite{vaswani2017attention} (see Fig.~\ref{fig:pipeline}) as the initial step in our feature encoding process. Specifically, within each range component $t_i \in \mathbf{T}$, which includes $N_r$ RT tokens, we utilize the Doppler bin descriptors as token features. Additionally, two index channels are employed for positional embeddings to enhance the specificity of our spatial encoding.
% The resulting sidelobe-repressed sparse RT is denoted as  $\mathbf{P}\in\mathbb{R}^{R\times N_r\times (8+2)}$. 

\noindent\textbf{Sequential sparse convolution.}  For efficiency, we apply a series of 3D sparse convolutions~\cite{yan2018second} onto the sparse RT for spatial feature encoding in the spherical voxel space. This produces a 3D dense feature volume $\mathbf{F}\in\mathbb{R}^{\frac{R}{S}\times \frac{A}{S}\times \frac{E}{S}\times C_f} (N_f = \frac{R}{S}\times \frac{A}{S}\times \frac{E}{S})$ with a reduce spatial dimension characterized by the stride $S$, where $C_f$ denotes the feature dimension. Note that $\mathbf{F}$ inherently aligns with the spherical space with each feature element's indices corresponding to a spherical coordinate. 

% \chris{there seems a notation mess for $C_f$, $K$ and $N_f$ in the above. Note we can delete the notations not used in the subsequent sections.}

\noindent\textbf{Deformable self-attention.} Following the consecutive 3D sparse convolution, we use the 3D deformable attention~\cite{zhu2020deformable} to further refine and augment our feature volume $\mathbf{F}$ by enforcing spatial interaction. As a definition, for a query feature $z$ corresponding to a reference point $p$ in the input feature $\mathbf{X}$, its feature can be updated by deformable attention in the following equation:
\begin{equation}
    \mathrm{DeformAttn} (z, p, \mathbf{X}) = \sum^{M}_{m=1}\mathbf{W}_m\left[\sum^{K}_{k=1}\mathbf{A}_{mk}\cdot\mathbf{W}'_m{\mathbf{X}}(p+\Delta p_{mk})\right]
\end{equation}
where $\mathbf{W}_m$ and $\mathbf{W}'_m$ are the learnable weights for the $m$-th attention head, while $\mathbf{A}_{mk}$ and $\Delta p_{mk}$ is the attention weight and sampling offset calculated with $z$ for its $k$-th sampling point and the $m$-th head. $\mathbf{X}(p+\Delta p_{mk})$ is the key features at the sample location $(p+\Delta p_{mk})$. By applying self-attention to $\mathbf{F}=\{f^q\}_{q=1}^{N_f}$, the refined feature volume $\mathbf{F}_r=\{f_r^q\}_{q=1}^{N_f}$ can be derived by:
\begin{equation}
    {f}_r^q = \mathrm{DeformAttn} (f^q, p^q, \mathbf{F})
\end{equation}

\subsection{Spherical-to-Cartesian feature aggregation}\label{aggregation}
Decoding 3D Cartesian occupancy from a spherical feature volume is inherently challenging due to misalignments in spatial axes and discrepancies in the regions they represent. An intuitive approach would be to transform the spherical feature volume into a Cartesian one and then decode the 3D Cartesian occupancy. However, this method can introduce feature-level interpolation errors, which we aim to avoid as discussed in Section~\ref{spherical}.

To avoid conducting interpolation, we propose to aggregate the spherical features in a \textit{learnable} way, with 3D volume queries defined in the Cartesian coordinates attending to the feature samples in $\mathbf{F}_r$, as shown in Fig.~\ref{fig:pipeline}. First, we build learnable grid-based voxel queries $\mathbf{H}=\{h^q\in\mathbb{R}^{C_f}\}_{q=1}^{H\times W\times L}$ which has the same volumetric size as our desired output $\mathbf{O}$ and the same feature dimension as the spherical feature volume $\mathbf{F}_r$. Each voxel query $h^q$ corresponds to a 3D point $p^q$ in the Cartesian coordinate. Second, the 3D point $p^q$ of each query is transformed from the Cartesian to the spherical coordinate, which is then mapped to a index position in $\mathbf{F}_r$ denoted as $\Phi (p^q)$. We take $\Phi (p^q)$ as a 3D reference point in the spherical space and sample key elements in its vicinity from the feature volume $\mathbf{F}_r$. Lastly, we leverage deformable cross-attention~\cite{zhu2020deformable} to aggregate the key samples for each reference point and the output $\mathbf{G}=\{g^q\in\mathbb{R}^{C_f}\}_{q=1}^{H\times W\times L}$ can be calculated by: 
\begin{equation}
    g^q = \mathrm{DeformAttn} (h^q, \Phi(p^q), \mathbf{F}_r)
\end{equation}

\subsection{3D occupancy decoding and supervision}\label{decoding}
With the aggregated voxel features $\mathbf{G}$, we leverage consecutive 3D convolutions~\cite{tran2015learning,zhou2018voxelnet} with skip connection~\cite{he2016deep} to decode hierarchical feature volumes at $N_s$ scales with a scaling step of 2. Multi-scale feature volumes are then merged in a top-down way~\cite{lin2017feature} via upsampling features by a factor 2 and concatenated along the feature dimension, resulting in $\mathbf{G}_d\in\mathbb{R}^{H\times W\times L\times N_sC_f}$. Finally, the occupancy head equipped with the \emph{softmax} function is employed to output the normalized occupancy probabilities $\tilde{\mathbf{O}}\in\{0,1\}^{H\times W\times L\times (C+1)}$ for all voxels on $C$ semantic classes and one free class.
% \subsection{Training loss}\label{loss}

Our network is trained in a supervised way with the ground truth occupancy. Following~\cite{wang2023openoccupancy}, we use the cross-entropy loss as the primary loss to optimize the training and incorporate the lovasz-softmax loss~\cite{berman2018lovasz} to handle the class imbalances. Moreover, we utilize the scene- and class-wise affinity loss proposed in~\cite{cao2022monoscene} to enhance the optimization of geometry and semantic IoU metrics.

\section{Experiment}

\subsection{Experimental setup}

\noindent\textbf{Dataset preparation.} Our experiments are conducted on the K-Radar dataset~\cite{paek2022k}, which is, to the best of our knowledge, the only autonomous driving dataset providing available 4DRT data. Besides, K-Radar also contains multi-modal data from LiDAR, camera, GPS-RTK and annotated 3D bounding boxes and tracking IDs, enabling us to compare between different modalities and generate 3D occupancy labels. Following~\cite{wei2023surroundocc,wang2023openoccupancy,li2023sscbench}, we generate occupancy ground truth by superimposing consecutive LiDAR sweeps and construct the dense 3D occupancy grids via voxelization. To handle scene dynamics, we register objects with the same tracking IDs across the sequence. As K-Radar does not annotate fine-grained point-level semantics, we segment the scene into the foreground (\emph{e.g.,} sedan, truck, pedestrian) and background using bounding boxes and label the voxel grids into three classes, including foreground, background and free. Many sequences in K-Radar were collected under adverse weather (\emph{i.e.,} sleet, rain, and snow), which results in non-negligible noise to the generated occupancy labels based on LiDAR sweeps. Therefore, we reserve this adverse-weather test split for qualitative comparison and only generate the occupancy labels for the well-condition sequences, which are separated into the training, validation and test splits.

\noindent\textbf{Evaluation protocol.} As the pioneering study of 3D occupancy prediction using the K-Radar dataset, we have tailored the evaluation protocol to align with our experimental needs. We define the Region of Interest (RoI) with specific dimensions: a front range of [0, 51.2m], a side range of [-25.6m, 25.6m], and a height range of [-2.6m, 3m]. The voxel resolution is set at 0.4m, resulting in a target occupancy volume of $128 \times 128 \times 14$ voxels. Consistent with established methods in the field~\cite{behley2019semantickitti,wang2023openoccupancy,li2023sscbench}, we employ the Intersection over Union (IoU) metric to evaluate the geometric accuracy of our occupancy predictions, focusing solely on the occupied or free status without integrating semantics. Additionally, to gauge the effectiveness of our foreground-background segmentation, we calculate the mean IoU (mIoU) across these two classes. In line with previous studies~\cite{li2023voxformer,li2023sscbench}, we present our findings across multiple ranges, specifically at 51.2m, 25.6m, and 12.8m.

\noindent\textbf{Competing methods.}  We benchmark {\sysname} against state-of-the-art methods employing different modalities. Given that recent studies do not use radar data for 3D occupancy prediction, we adapt the OpenOccupancy LiDAR-based baseline and CONet~\cite{wang2023openoccupancy} to accommodate radar point cloud (RPC) inputs for our comparative analysis. Furthermore, we convert 4DRTs to Cartesian coordinates~\cite{paek2022k} with a voxel size of 0.4m, referred to as 4DRT-XYZ, and integrate them into the LiDAR-based OpenOccupancy framework~\cite{wang2023openoccupancy}. Following best practices from~\cite{paek2022k,paek2023enhanced}, we process 4DRT-XYZ into a sparser format. For a comprehensive inter-modality evaluation, we also replicate the OpenOccupancy LiDAR-based baseline~\cite{wang2023openoccupancy} and both monocular and stereo camera-based SurroundOcc~\cite{wei2023surroundocc} configurations to fit our experimental setup. Notably, we enrich our comparisons by generating 16-beam and 32-beam LiDAR point clouds from the standard 64-beam configurations through elevation-wise downsampling. The evaluation focuses on the overlap area between the horizontal field of view (FoV) of all sensors and our defined RoI to minimize potential data discrepancies beyond the FoV. For implementation, we train all evaluated models on our K-Radar well-condition training set.
% For fairness, camera-based methods utilize the monocular front-view image alone while LiDAR-based approaches use only the front half of each sweep.
% In particular, we calculate the intersection of the field of view (FoV) from all sensors and only compare the performance within this area. 

\begin{table*}[!tb]\small
    \renewcommand\arraystretch{1}
    \setlength\tabcolsep{3pt}
    \centering
    \resizebox{\textwidth}{!}{
    \begin{tabular}{ll|ccc|ccc|ccc|ccc}
        \toprule
        & & \multicolumn{3}{c|}{IoU (\%)} & \multicolumn{3}{c|}{mIoU (\%)} & \multicolumn{3}{c|}{\tikz \fill [bg] (0,0) rectangle (0.6em,0.6em); BG IoU (\%)} & \multicolumn{3}{c}{\tikz \fill [fg] (0,0) rectangle (0.6em,0.6em); FG IoU (\%)} \\
        \midrule
        Method & Input & 12.8m & 25.6m & 51.2m & 12.8m & 25.6m & 51.2m & 12.8m & 25.6m & 51.2m & 12.8m & 25.6m & 51.2m \\
        \midrule 
        {L-baseline~\cite{wang2023openoccupancy}} & RPC & 42.8 & 34.9 & 27.9 & 23.5 & 18.6 & 14.6 & 43.5 & 34.6 & 27.3 & 3.5 & 2.6 & 1.9 \\
        {L-CONet~\cite{wang2023openoccupancy}} & RPC & 46.1 & 36.0 & 25.0 & 24.6 & 20.3 & 14.4 & 43.3 & 35.4 & 25.6 & 5.8 & 5.2 & 3.1 \\
        {L-baseline~\cite{wang2023openoccupancy}} & 4DRT-XYZ & 47.4 & 38.1 & 28.5 & 29.9 & 24.3 & 17.5 & 46.4 & 37.5 & 27.9 & 13.4 & 11.1 & 7.2 \\
        \midrule
        RadarOcc (Ours) & 4DRT & \textbf{48.8} & \textbf{39.1} & \textbf{30.4} & \textbf{34.3} & \textbf{28.5} & \textbf{22.6} & \textbf{47.9} & \textbf{38.2} & \textbf{29.4} & \textbf{20.7} & \textbf{18.7} & \textbf{15.8} \\
        \bottomrule
    \end{tabular}
    }
    \caption{Quantitative comparison between {\sysname} and state-of-the-art radar-based baseline methods. Results are reported on K-Radar well-condition test split. Best result is shown in \textbf{bold}.}
    \label{tab:radar}
\end{table*}

\begin{table*}[!tb]\small
    \renewcommand\arraystretch{1}
    \setlength\tabcolsep{4pt}
    \centering
    \resizebox{\textwidth}{!}{
    \begin{tabular}{l|l|ccc|ccc|ccc|ccc}
    \toprule
        & & \multicolumn{3}{c|}{IoU (\%)} & \multicolumn{3}{c|}{mIoU (\%)} & \multicolumn{3}{c|}{\tikz \fill [bg] (0,0) rectangle (0.6em,0.6em); BG IoU (\%)} & \multicolumn{3}{c}{\tikz \fill [fg] (0,0) rectangle (0.6em,0.6em); FG IoU (\%)} \\
    \midrule
        & Method & 12.8m & 25.6m & 51.2m & 12.8m & 25.6m & 51.2m & 12.8m & 25.6m & 51.2m & 12.8m & 25.6m & 51.2m \\
    \midrule
        (a) & Ours & \textbf{48.8} & 39.1 & \textbf{30.4} & \textbf{34.3} & 28.5 & \textbf{22.6} & \textbf{47.9} & 38.2 & \textbf{29.4} & \textbf{20.7} & 18.7 & 15.8 \\
        (b) & Ours w/o DBD & 48.1 & \textbf{39.4} & 30.0 & 33.6 & \textbf{28.9} & 22.6 & 47.2 & \textbf{38.7} & 29.2 & 20.0 & \textbf{19.1} & \textbf{16.0} \\
        (c) & Ours w/o SSS & 44.2& 36.8& 28.7& 24.1& 20.2& 15.6&42.3 &35.6 & 27.6&5.9 &4.7 &3.5 \\
        (d) & Ours w/o SFE & 46.2 & 38.4 & 29.4 & 30.4 & 26.5 & 21.1 & 45.5 & 37.5 & 28.5 & 15.4 & 15.5 & 13.9 \\
    \bottomrule
    \end{tabular}
    }
    \caption{Ablation studies on key designs of {\sysname}. DBD, SSS, SFE refer to the Doppler bins descriptor, sidelobe-aware spatial sparfiying, and spherical-based feature encoding, respectively.}
    \label{tab:ablation}
\end{table*}

\subsection{Comparison against radar-based methods}
We first compare {\sysname} with state-of-the-art baseline methods using radar data for 3D occupancy prediction in Tab.~\ref{tab:radar}. As can be seen, {\sysname} outperforms other approaches in every metric, demonstrating its state-of-the-art performance in radar-based 3D occupancy prediction. Specifically, our 4DRT-based {\sysname} largely improves the performance over RPC-based methods: the mIoU of L-CONet~\cite{wang2023openoccupancy} is relatively improved by 39.4\%, 40.4\% and 56.9\% for different volumes (12.8m, 25.6m, 51.2m). Such a significant improvement mainly stems from the dense data format of 4DRT, which retains critical information from low-reflectivity objects, enabling effective occupancy prediction for the whole scene. 4DRT-XYZ based L-baseline~\cite{wang2023openoccupancy} also outperforms RPC-based methods but inferior to {\sysname}, especially in long-range FG IoU. We credit this to the interpolation errors led to small and far foreground objects when we converting 4DRT to Cartesian coordinates. 

\subsection{Ablation study} \label{ablationstudy}
To validate the effectiveness of our key designs, we ablate them alone from our 4DRT-based pipeline {\sysname} and show the evaluation results on K-Radar well-condition test split in Tab.~\ref{tab:ablation}.

\noindent\textbf{Doppler bins descriptor.} By replacing the Doppler bins descriptor with the average-pooling result, the performance of {\sysname} is degraded in most metrics (row (a) vs. (b) in Tab.~\ref{tab:ablation}), demonstrating the usefulness of preserving the information encoded by the Doppler axis (\emph{c.f.} Sec.~\ref{reduction}). However, the improvement is somehow marginal due to the limited Doppler measurement range of the radar used in K-Radar~\cite{paek2022k}, which wraps around the overflow values, causing ambiguity in Doppler velocity.   

\noindent\textbf{Sidelobe-aware spatial sparsifying.} We conduct this experiment (row (c) in Tab.~\ref{tab:ablation}) by changing our sidelobe-aware spatial sparsifying (\emph{c.f.} Sec.~\ref{reduction}) to the percentile-based spatial sparsifying used in~\cite{paek2022k,paek2023enhanced}. Our sidelobe-aware approach leads to a remarkable advancement in performance, especially in mIoU metrics. This is attributed to its ability to preserve more valid elements from diverse ranges and suppress sidelobes for sparse RTs, allowing for more accurate prediciton.

\noindent\textbf{Spherical-based feature encoding.} For row (d) in Tab.~\ref{tab:ablation}, we transform sparse RT to Cartesian coordinates before feature encoding (\emph{c.f.} Sec.~\ref{spherical}) and omit the spherical-to-Cartesian feature aggregation (\emph{c.f.} Sec.~\ref{aggregation}). We can see that our spherical-based feature encoding gains the performance for each metric as our strategy preserves the original data distribution, avoiding incurring interpolation errors. This also validates the effectiveness of our learnable spherical-to-Cartesian feature aggregation.

\subsection{Model efficiency}
To assess the runtime efficiency of \sysname, we conducted our model inference on a single Nvidia GTX 3090 GPU. The results shows an average inference speed of approximately 3.3fps. Although there is still a gap between the real-time application (\emph{i.e.,} 10fps), our inference speed has surpassed that of many camera-based methods as reported in~\cite{wei2023surroundocc}. Further improvements in inference speed can be achieved by reducing network complexity and applying precision reduction techniques, such as converting model precision from \textit{Float32} (FP32) to \textit{Float16} (FP16). 

\begin{table*}[!tb]\small
    \renewcommand\arraystretch{1}
    \setlength\tabcolsep{2pt}
    \centering
    \resizebox{\textwidth}{!}{
    \begin{tabular}{l|c|c|c|c|c|c|c}
        \toprule
        Method & range-wise attn. & seq. sparse conv. & deform. self-attn. & deform. cross-attn. & occ. decoding & total runtime & fps  \\
        \midrule 
        RadarOcc & 2.5 & 47.5 & 88.8 & 72.0 & 92.1 & 302.9 & 3.30  \\
        RadarOcc (w. optim.) & 2.5 & 20.7 (-56.4\%) & 32.8 (-63.1\%) & 29.7 (-58.7\%) & 48.3 (-47.6\%) & 133.9 (-55.8\%) & 7.46 (+126.1\%)\\
        \bottomrule
        % Component & range-wise attn. & seq. sparse conv. &  \\
    \end{tabular}
    }
    \caption{Comparison between {\sysname} and its lightweight version after computation optimization in terms of each component’s and total runtime (ms) and fps. Relative change is shown in ($\cdot$).}
    \label{tab:speed}
    \vspace{-1em}
\end{table*}

\begin{table*}[!tb]\small
    \renewcommand\arraystretch{1}
    \setlength\tabcolsep{6pt}
    \centering
    \resizebox{\textwidth}{!}{
    \begin{tabular}{l|ccc|ccc|ccc|ccc}
        \toprule
        & \multicolumn{3}{c|}{IoU (\%)} & \multicolumn{3}{c|}{mIoU (\%)} & \multicolumn{3}{c|}{\tikz \fill [bg] (0,0) rectangle (0.6em,0.6em); BG IoU (\%)} & \multicolumn{3}{c}{\tikz \fill [fg] (0,0) rectangle (0.6em,0.6em); FG IoU (\%)} \\
        \midrule
        Method & 12.8m & 25.6m & 51.2m & 12.8m & 25.6m & 51.2m & 12.8m & 25.6m & 51.2m & 12.8m & 25.6m & 51.2m \\
        \midrule 
        RadarOcc &  \textbf{48.8} & \textbf{39.1} & \textbf{30.4} & {34.3} & \textbf{28.5} & \textbf{22.6} & \textbf{47.9} & \textbf{38.2} & \textbf{29.4} & {20.7} & \textbf{18.7} & \textbf{15.8} \\
        RadarOcc (w. optim.) &  46.5 & 38.0 & 29.3 & \textbf{35.5} & 27.6 & 20.9 & 46.0 & 37.6 & 28.8 & \textbf{25.0} & 17.5 & 13.1\\
        \bottomrule
    \end{tabular}
    }
    \caption{Comparison between {\sysname} and its lightweight version after computation optimization in terms of performance across metrics at different ranges. Better result is shown in \textbf{bold}.}
    \label{tab:performance}
    \vspace{-1em}
\end{table*}

\begin{table*}[!tb]\small
    \renewcommand\arraystretch{1}
    \setlength\tabcolsep{4pt}
    \centering
    \resizebox{\textwidth}{!}{

    \begin{tabular}{ll|ccc|ccc|ccc|ccc}
        \toprule
        & & \multicolumn{3}{c|}{IoU (\%)} & \multicolumn{3}{c|}{mIoU (\%)} & \multicolumn{3}{c|}{\tikz \fill [bg] (0,0) rectangle (0.6em,0.6em); BG IoU (\%)} & \multicolumn{3}{c}{\tikz \fill [fg] (0,0) rectangle (0.6em,0.6em); FG IoU (\%)} \\
        \midrule
        Method & Input & 12.8m & 25.6m & 51.2m & 12.8m & 25.6m & 51.2m & 12.8m & 25.6m & 51.2m & 12.8m & 25.6m & 51.2m \\
        \midrule 
        \multirow{3}{*}{L-baseline~\cite{wang2023openoccupancy}} & 
        L (16) & \textbf{\textcolor{myblue}{49.1}} & \textbf{\textcolor{myblue}{43.3}} & \textbf{\textcolor{mygreen}{35.2}} & \textbf{\textcolor{myorange}{39.0}} & \textbf{\textcolor{myblue}{34.3}} & \textbf{\textcolor{myblue}{28.2}} & \textbf{\textcolor{myblue}{48.2}} & \textbf{\textcolor{myblue}{42.5}} & \textbf{\textcolor{mygreen}{34.4}} & \textbf{\textcolor{myorange}{29.8}} & \textbf{\textcolor{myblue}{26.1}} & \textbf{\textcolor{myblue}{22.1}} \\
        & L (32) & \textbf{\textcolor{mygreen}{51.1}} & \textbf{\textcolor{mygreen}{44.0}} & \textbf{\textcolor{myblue}{34.9}} & \textbf{\textcolor{mygreen}{42.1}} & \textbf{\textcolor{mygreen}{35.0}} & \textbf{\textcolor{mygreen}{28.9}} & \textbf{\textcolor{mygreen}{50.8}} & \textbf{\textcolor{mygreen}{43.6}} & \textbf{\textcolor{myblue}{34.2}} & \textbf{\textcolor{mygreen}{33.5}} & \textbf{\textcolor{mygreen}{26.3}} & \textbf{\textcolor{mygreen}{23.6}} \\ 
        & L (64) & \textbf{\textcolor{myred}{56.9}} & \textbf{\textcolor{myred}{52.5}} & \textbf{\textcolor{myred}{43.8}} & \textbf{\textcolor{myred}{53.7}} & \textbf{\textcolor{myred}{45.2}} & \textbf{\textcolor{myred}{36.6}} & \textbf{\textcolor{myred}{56.1}} & \textbf{\textcolor{myred}{51.8}} & \textbf{\textcolor{myred}{43.3}} & \textbf{\textcolor{myred}{51.2}} & \textbf{\textcolor{myred}{36.5}} & \textbf{\textcolor{myred}{29.9}} \\
        \midrule
        \multirow{2}{*}{SurroundOcc~\cite{wei2023surroundocc}} & C & 44.3 & 33.1 & 24.1 & 36.1 & 23.9 & 14.7 & 44.1 & 32.9 & 23.7 & 28.2 & 15.0 & 5.7\\
        & C (S) & {46.2} & 34.4 & 25.4 & \textbf{\textcolor{myblue}{40.8}} & 25.4 & 16.2 & {{45.5}} & 34.1 & 25.1 & \textbf{\textcolor{myblue}{36.1}} & 16.7 & 7.3\\
        \midrule
        RadarOcc (Ours) & 4DRT & \textbf{\textcolor{myorange}{48.8}} & \textbf{\textcolor{myorange}{39.1}} & \textbf{\textcolor{myorange}{30.4}} & {34.3} & \textbf{\textcolor{myorange}{28.5}} & \textbf{\textcolor{myorange}{22.6}} & \textbf{\textcolor{myorange}{47.9}} & \textbf{\textcolor{myorange}{38.2}} & \textbf{\textcolor{myorange}{29.4}} & {20.7} & \textbf{\textcolor{myorange}{18.7}} & \textbf{\textcolor{myorange}{15.8}} \\
        \bottomrule
        \end{tabular}
    }  
    \caption{Quantitative comparison between {\sysname} and state-of-the-art methods based on LiDAR and camera. Results are reported on K-Radar well-condition test split. $(\cdot)$ is the number of LiDAR beams and (S) denotes stereo. The top four methods are colored as \textbf{\textcolor{myred}{red}}, \textbf{\textcolor{mygreen}{green}}, \textbf{\textcolor{myblue}{blue}}, and \textbf{\textcolor{myorange}{orange}}.}
    \label{tab:results}
\end{table*}

To validate this, we simplified the feature encoding (\emph{c.f.} Sec.~\ref{spherical}) and aggregation (\emph{c.f.} Sec.~\ref{aggregation}) modules by reducing some redundancy layer (\emph{e.g.}, number of layers in deformable attention) for efficiency, and converted the computationally intensive 3D occupancy decoding module (\emph{c.f.} Sec.~\ref{decoding}) from FP32 to FP16 via the quantization in PyTorch. These optimizations resulted in a 126\% increase in inference speed, reaching approximately 7.46 fps, with only a minimal impact on performance. Please refer to Tab.~\ref{tab:speed} and Tab.~\ref{tab:performance} for detailed changes in runtime for each module and performance. Given the increasing computational power of modern embedded GPUs, such as the Nvidia Jetson Orin, which can almost rival desktop GPUs like the Nvidia GTX 2090, we believe this enhanced inference speed demonstrates the potential for real-time application of our method in future vehicle systems, especially if further model quantization is applied. 

\subsection{Comparison between different modalities}
\begin{figure}[!tbp]
    \centering
    \includegraphics[width=\textwidth]{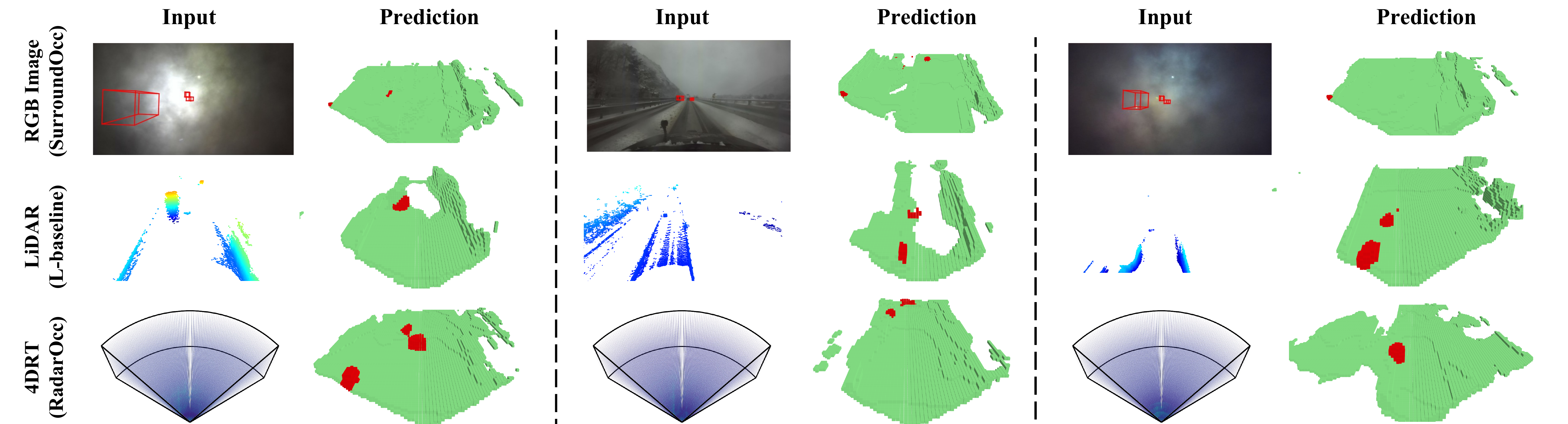}
    % \fbox{\rule[-.5cm]{0cm}{5cm} \rule[-.5cm]{5cm}{0cm}}
    \caption{Qualitative comparison between {\sysname}, LiDAR-based L-baseline~\cite{wang2023openoccupancy} and camera-based SurroundOcc~\cite{wei2023surroundocc} in adverse weathers.
    Ground truth bounding boxes are shown in RGB images.}
    \label{fig:adverse}
\end{figure}

To enrich our benchmark results and provide insights into the performance comparison between different modalities, we also evaluate state-of-the-art baseline methods~\cite{wang2023openoccupancy,wei2023surroundocc} on LiDAR and camera input. Quantitative results on K-Radar well-condition test split are reported in Tab.~\ref{tab:results}, while examples of qualitative results on K-Radar adverse-weather testing splits are exhibited in Fig.~\ref{fig:adverse}.

\noindent\textbf{Quantitative results under normal weathers.} As seen in Tab.~\ref{tab:results}, not surprisingly, LiDAR-based L-baselines~\cite{wang2023openoccupancy} rank the top three in most metrics thanks to LiDAR's low-noise and high-resolution measurements (\emph{vs.} radar) and direct depth measurement (\emph{vs.} camera). Due to the inherently lower resolution and considerable noise of radar data, radar-based methods exhibit inferior to LiDAR-based methods in normal weather. However, {\sysname} still shows comparable performance to 16-beam LiDAR, and surpasses monocular and stereo camera-based method in most metrics. Notably, {\sysname} outperforms state-of-the-art SurroundOcc~\cite{wei2023surroundocc} relatively by 39.5\%/19.7\% and 53.7\%/26.1\% in mIoU/IoU@51.2m for stereo and monocular input, respectively. Stereo camera-based SurroundOcc~\cite{wei2023surroundocc} ranks third on FG IoU and mIoU@12.8m because of stereo vision's ability to infer accurate depth at short ranges, where the disparity between the two images is more pronounced. 

\noindent\textbf{Qualitative results under adverse weathers.} While we have demonstrated the competitive performance of {\sysname} under normal weather, the key reason behind using radar for perception comes from its unique robustness against adverse weather where LiDAR and cameras fall short. To showcase such an inherent advantage, we provide some examples of qualitative results from different modalities in Fig.~\ref{fig:adverse}. As can be seen, {\sysname} provide robust  3D occupancy prediction under heavy rain and snow. In contrast, the camera lens are covered by the rain/snow and LiDAR measurements of some objects ahead are missing as water droplets or snowflakes can scatter and absorb the laser beams, leading to worse results. Please see our supplementary materials for more qualitative results.

\section{Conclusion}
In this work, we propose \sysname, a novel 3D occupancy prediction approach based on 4DRTs output from 4D imaging radar, enabling robust all-weather perception for autonomous vehicles. We analyse the rationale and challenges of using 4DRTs for 3D occupancy prediction and present tailored solutions to cope with the large, noisy and spherical 4DRTs. Experiments on the K-Radar dataset show {\sysname}'s state-of-the-art performance in radar-based 3D occupancy prediction and comparable results to other modalities in normal weathers. Through qualitative analysis, we also exhibit its unique robustness against various adverse weathers. We believe our work could endorse the potential of 4D imaging radar to be an alternative to LiDAR and setup an effective baseline for further research and development of 4D radar-based occupancy perception.

\noindent\textbf{Limitations.} As an initial investigation into 4D radar-based 3D occupancy prediction, this work has several limitations as follows. First, our method maps single-frame 4D radar data to single-frame 3D occupancy prediction without modeling the temporal information and performing occupancy forecasting. Second, due to the lack of point-wise annotation, our task is limited to two general semantics, \emph{i.e.,} foreground and background. Future work will aim to address these issues.
% \section{Submission of papers to NeurIPS 2024}

\section*{Acknowledgement}
This research is partially supported by the Engineering and Physical Sciences Research Council (EPSRC) under the Centre for Doctoral Training in Robotics and Autonomous Systems at the Edinburgh Centre of Robotics (EP/S023208/1).

\bibliographystyle{unsrt}
\bibliography{egbib}

\begin{thebibliography}{100}

\bibitem{youtube2022tesla}
Tesla.
\newblock {Tesla AI Day 2022}.
\newblock \url{https://www.youtube.com/watch?v=ODSJsviD_SU}, 2022.
\newblock Accessed: 2024-04-08.

\bibitem{agro2023implicit}
Ben Agro, Quinlan Sykora, Sergio Casas, and Raquel Urtasun.
\newblock Implicit occupancy flow fields for perception and prediction in self-driving.
\newblock In {\em Proceedings of the IEEE/CVF Conference on Computer Vision and Pattern Recognition (CVPR)}, pages 1379--1388, 2023.

\bibitem{wang2023openoccupancy}
Xiaofeng Wang, Zheng Zhu, Wenbo Xu, Yunpeng Zhang, Yi~Wei, Xu~Chi, Yun Ye, Dalong Du, Jiwen Lu, and Xingang Wang.
\newblock Openoccupancy: A large scale benchmark for surrounding semantic occupancy perception.
\newblock In {\em Proceedings of the IEEE/CVF International Conference on Computer Vision (ICCV)}, pages 17850--17859, 2023.

\bibitem{tian2024occ3d}
Xiaoyu Tian, Tao Jiang, Longfei Yun, Yucheng Mao, Huitong Yang, Yue Wang, Yilun Wang, and Hang Zhao.
\newblock Occ3d: A large-scale 3d occupancy prediction benchmark for autonomous driving.
\newblock {\em Advances in Neural Information Processing Systems}, 36, 2024.

\bibitem{li2023voxformer}
Yiming Li, Zhiding Yu, Christopher Choy, Chaowei Xiao, Jose~M Alvarez, Sanja Fidler, Chen Feng, and Anima Anandkumar.
\newblock Voxformer: Sparse voxel transformer for camera-based 3d semantic scene completion.
\newblock In {\em Proceedings of the IEEE/CVF Conference on Computer Vision and Pattern Recognition (CVPR)}, pages 9087--9098, 2023.

\bibitem{liang2022bevfusion}
Tingting Liang, Hongwei Xie, Kaicheng Yu, Zhongyu Xia, Zhiwei Lin, Yongtao Wang, Tao Tang, Bing Wang, and Zhi Tang.
\newblock Bevfusion: A simple and robust lidar-camera fusion framework.
\newblock {\em Advances in Neural Information Processing Systems}, 35:10421--10434, 2022.

\bibitem{liu2023bevfusion}
Zhijian Liu, Haotian Tang, Alexander Amini, Xinyu Yang, Huizi Mao, Daniela~L Rus, and Song Han.
\newblock Bevfusion: Multi-task multi-sensor fusion with unified bird's-eye view representation.
\newblock In {\em Proceedings of the IEEE International Conference on Robotics and Automation (ICRA)}, pages 2774--2781. IEEE, 2023.

\bibitem{chen2023futr3d}
Xuanyao Chen, Tianyuan Zhang, Yue Wang, Yilun Wang, and Hang Zhao.
\newblock Futr3d: A unified sensor fusion framework for 3d detection.
\newblock In {\em Proceedings of the IEEE/CVF Conference on Computer Vision and Pattern Recognition (CVPR)}, pages 172--181, 2023.

\bibitem{kong2023rethinking}
Lingdong Kong, Youquan Liu, Runnan Chen, Yuexin Ma, Xinge Zhu, Yikang Li, Yuenan Hou, Yu~Qiao, and Ziwei Liu.
\newblock Rethinking range view representation for lidar segmentation.
\newblock In {\em Proceedings of the IEEE/CVF International Conference on Computer Vision (ICCV)}, pages 228--240, 2023.

\bibitem{chen2021polarstream}
Qi~Chen, Sourabh Vora, and Oscar Beijbom.
\newblock Polarstream: Streaming object detection and segmentation with polar pillars.
\newblock {\em Advances in Neural Information Processing Systems}, 34:26871--26883, 2021.

\bibitem{lai2022stratified}
Xin Lai, Jianhui Liu, Li~Jiang, Liwei Wang, Hengshuang Zhao, Shu Liu, Xiaojuan Qi, and Jiaya Jia.
\newblock Stratified transformer for 3d point cloud segmentation.
\newblock In {\em Proceedings of the IEEE/CVF Conference on Computer Vision and Pattern Recognition (CVPR)}, pages 8500--8509, 2022.

\bibitem{liang2020pnpnet}
Ming Liang, Bin Yang, Wenyuan Zeng, Yun Chen, Rui Hu, Sergio Casas, and Raquel Urtasun.
\newblock Pnpnet: End-to-end perception and prediction with tracking in the loop.
\newblock In {\em Proceedings of the IEEE/CVF Conference on Computer Vision and Pattern Recognition (CVPR)}, pages 11553--11562, 2020.

\bibitem{gu2023vip3d}
Junru Gu, Chenxu Hu, Tianyuan Zhang, Xuanyao Chen, Yilun Wang, Yue Wang, and Hang Zhao.
\newblock Vip3d: End-to-end visual trajectory prediction via 3d agent queries.
\newblock In {\em Proceedings of the IEEE/CVF Conference on Computer Vision and Pattern Recognition (CVPR)}, pages 5496--5506, 2023.

\bibitem{hu2023planning}
Yihan Hu, Jiazhi Yang, Li~Chen, Keyu Li, Chonghao Sima, Xizhou Zhu, Siqi Chai, Senyao Du, Tianwei Lin, Wenhai Wang, et~al.
\newblock Planning-oriented autonomous driving.
\newblock In {\em Proceedings of the IEEE/CVF Conference on Computer Vision and Pattern Recognition (CVPR)}, pages 17853--17862, 2023.

\bibitem{cheng2021s3cnet}
Ran Cheng, Christopher Agia, Yuan Ren, Xinhai Li, and Liu Bingbing.
\newblock S3cnet: A sparse semantic scene completion network for lidar point clouds.
\newblock In {\em Proceedings of the Conference on Robot Learning (CoRL)}, pages 2148--2161. PMLR, 2021.

\bibitem{roldao2020lmscnet}
Luis Roldao, Raoul de~Charette, and Anne Verroust-Blondet.
\newblock Lmscnet: Lightweight multiscale 3d semantic completion.
\newblock In {\em Proceedings of the International Conference on 3D Vision (3DV)}, pages 111--119, 2020.

\bibitem{yan2021sparse}
Xu~Yan, Jiantao Gao, Jie Li, Ruimao Zhang, Zhen Li, Rui Huang, and Shuguang Cui.
\newblock Sparse single sweep lidar point cloud segmentation via learning contextual shape priors from scene completion.
\newblock In {\em Proceedings of the AAAI Conference on Artificial Intelligence (AAAI)}, volume~35, pages 3101--3109, 2021.

\bibitem{li2021semi}
Pengfei Li, Yongliang Shi, Tianyu Liu, Hao Zhao, Guyue Zhou, and Ya-Qin Zhang.
\newblock Semi-supervised implicit scene completion from sparse lidar.
\newblock {\em arXiv preprint arXiv:2111.14798}, 2021.

\bibitem{rist2021semantic}
Christoph~B Rist, David Emmerichs, Markus Enzweiler, and Dariu~M Gavrila.
\newblock Semantic scene completion using local deep implicit functions on lidar data.
\newblock {\em IEEE Transactions on Pattern Analysis and Machine Intelligence}, 44(10):7205--7218, 2021.

\bibitem{liu2023lidar}
Xinhao Liu, Moonjun Gong, Qi~Fang, Haoyu Xie, Yiming Li, Hang Zhao, and Chen Feng.
\newblock Lidar-based 4d occupancy completion and forecasting.
\newblock {\em arXiv preprint arXiv:2310.11239}, 2023.

\bibitem{xia2023scpnet}
Zhaoyang Xia, Youquan Liu, Xin Li, Xinge Zhu, Yuexin Ma, Yikang Li, Yuenan Hou, and Yu~Qiao.
\newblock Scpnet: Semantic scene completion on point cloud.
\newblock In {\em Proceedings of the IEEE/CVF Conference on Computer Vision and Pattern Recognition}, pages 17642--17651, 2023.

\bibitem{khurana2023point}
Tarasha Khurana, Peiyun Hu, David Held, and Deva Ramanan.
\newblock Point cloud forecasting as a proxy for 4d occupancy forecasting.
\newblock In {\em Proceedings of the IEEE/CVF Conference on Computer Vision and Pattern Recognition (CVPR)}, pages 1116--1124, 2023.

\bibitem{cao2022monoscene}
Anh-Quan Cao and Raoul De~Charette.
\newblock Monoscene: Monocular 3d semantic scene completion.
\newblock In {\em Proceedings of the IEEE/CVF Conference on Computer Vision and Pattern Recognition (CVPR)}, pages 3991--4001, 2022.

\bibitem{huang2023tri}
Yuanhui Huang, Wenzhao Zheng, Yunpeng Zhang, Jie Zhou, and Jiwen Lu.
\newblock Tri-perspective view for vision-based 3d semantic occupancy prediction.
\newblock In {\em Proceedings of the IEEE/CVF Conference on Computer Vision and Pattern Recognition (CVPR)}, pages 9223--9232, 2023.

\bibitem{wei2023surroundocc}
Yi~Wei, Linqing Zhao, Wenzhao Zheng, Zheng Zhu, Jie Zhou, and Jiwen Lu.
\newblock Surroundocc: Multi-camera 3d occupancy prediction for autonomous driving.
\newblock In {\em Proceedings of the IEEE/CVF International Conference on Computer Vision (ICCV)}, pages 21729--21740, 2023.

\bibitem{tong2023scene}
Wenwen Tong, Chonghao Sima, Tai Wang, Li~Chen, Silei Wu, Hanming Deng, Yi~Gu, Lewei Lu, Ping Luo, Dahua Lin, et~al.
\newblock Scene as occupancy.
\newblock In {\em Proceedings of the IEEE/CVF International Conference on Computer Vision (ICCV)}, pages 8406--8415, 2023.

\bibitem{zhang2023occformer}
Yunpeng Zhang, Zheng Zhu, and Dalong Du.
\newblock Occformer: Dual-path transformer for vision-based 3d semantic occupancy prediction.
\newblock In {\em Proceedings of the IEEE/CVF International Conference on Computer Vision (ICCV)}, pages 9433--9443, 2023.

\bibitem{tan2023ovo}
Zhiyu Tan, Zichao Dong, Cheng Zhang, Weikun Zhang, Hang Ji, and Hao Li.
\newblock Ovo: Open-vocabulary occupancy.
\newblock {\em arXiv preprint arXiv:2305.16133}, 2023.

\bibitem{huang2023selfocc}
Yuanhui Huang, Wenzhao Zheng, Borui Zhang, Jie Zhou, and Jiwen Lu.
\newblock Selfocc: Self-supervised vision-based 3d occupancy prediction.
\newblock {\em arXiv preprint arXiv:2311.12754}, 2023.

\bibitem{zhang2023occnerf}
Chubin Zhang, Juncheng Yan, Yi~Wei, Jiaxin Li, Li~Liu, Yansong Tang, Yueqi Duan, and Jiwen Lu.
\newblock Occnerf: Self-supervised multi-camera occupancy prediction with neural radiance fields.
\newblock {\em arXiv preprint arXiv:2312.09243}, 2023.

\bibitem{vobecky2024pop}
Antonin Vobecky, Oriane Sim{\'e}oni, David Hurych, Spyridon Gidaris, Andrei Bursuc, Patrick P{\'e}rez, and Josef Sivic.
\newblock Pop-3d: Open-vocabulary 3d occupancy prediction from images.
\newblock {\em Advances in Neural Information Processing Systems}, 36, 2024.

\bibitem{ma2023cotr}
Qihang Ma, Xin Tan, Yanyun Qu, Lizhuang Ma, Zhizhong Zhang, and Yuan Xie.
\newblock Cotr: Compact occupancy transformer for vision-based 3d occupancy prediction.
\newblock {\em arXiv preprint arXiv:2312.01919}, 2023.

\bibitem{ma2023cam4docc}
Junyi Ma, Xieyuanli Chen, Jiawei Huang, Jingyi Xu, Zhen Luo, Jintao Xu, Weihao Gu, Rui Ai, and Hesheng Wang.
\newblock Cam4docc: Benchmark for camera-only 4d occupancy forecasting in autonomous driving applications.
\newblock {\em arXiv preprint arXiv:2311.17663}, 2023.

\bibitem{sun20214d}
Shunqiao Sun and Yimin~D Zhang.
\newblock 4d automotive radar sensing for autonomous vehicles: A sparsity-oriented approach.
\newblock {\em IEEE Journal of Selected Topics in Signal Processing}, 15(4):879--891, 2021.

\bibitem{sun2020mimo}
Shunqiao Sun, Athina~P Petropulu, and H~Vincent Poor.
\newblock Mimo radar for advanced driver-assistance systems and autonomous driving: Advantages and challenges.
\newblock {\em IEEE Signal Processing Magazine}, 37(4):98--117, 2020.

\bibitem{meyer2019automotive}
Michael Meyer and Georg Kuschk.
\newblock Automotive radar dataset for deep learning based 3d object detection.
\newblock In {\em Proceedings of the 16th European radar conference (EuRAD)}, pages 129--132. IEEE, 2019.

\bibitem{liu2024echoes}
Yang Liu, Feng Wang, Naiyan Wang, and ZHAO-XIANG ZHANG.
\newblock Echoes beyond points: Unleashing the power of raw radar data in multi-modality fusion.
\newblock {\em Advances in Neural Information Processing Systems}, 36, 2024.

\bibitem{Rebut_2022_CVPR}
Julien Rebut, Arthur Ouaknine, Waqas Malik, and Patrick P\'erez.
\newblock Raw high-definition radar for multi-task learning.
\newblock In {\em Proceedings of the IEEE/CVF Conference on Computer Vision and Pattern Recognition (CVPR)}, pages 17021--17030, June 2022.

\bibitem{xu2021rpfa}
Baowei Xu, Xinyu Zhang, Li~Wang, Xiaomei Hu, Zhiwei Li, Shuyue Pan, Jun Li, and Yongqiang Deng.
\newblock Rpfa-net: A 4d radar pillar feature attention network for 3d object detection.
\newblock In {\em Proceedings of the IEEE International Intelligent Transportation Systems Conference (ITSC)}, pages 3061--3066. IEEE, 2021.

\bibitem{palffy2022multi}
Andras Palffy, Ewoud Pool, Srimannarayana Baratam, Julian~FP Kooij, and Dariu~M Gavrila.
\newblock Multi-class road user detection with 3+ 1d radar in the view-of-delft dataset.
\newblock {\em IEEE Robotics and Automation Letters}, 7(2):4961--4968, 2022.

\bibitem{tan20223d}
Bin Tan, Zhixiong Ma, Xichan Zhu, Sen Li, Lianqing Zheng, Sihan Chen, Libo Huang, and Jie Bai.
\newblock 3d object detection for multi-frame 4d automotive millimeter-wave radar point cloud.
\newblock {\em IEEE Sensors Journal}, 2022.

\bibitem{paek2022k}
Dong-Hee Paek, Seung-Hyun Kong, and Kevin~Tirta Wijaya.
\newblock K-radar: 4d radar object detection for autonomous driving in various weather conditions.
\newblock {\em Advances in Neural Information Processing Systems}, 35:3819--3829, 2022.

\bibitem{meyer2019deep}
Michael Meyer and Georg Kuschk.
\newblock Deep learning based 3d object detection for automotive radar and camera.
\newblock In {\em Proceedings of the 16th European Radar Conference (EuRAD)}, pages 133--136. IEEE, 2019.

\bibitem{wang2022interfusion}
Li~Wang, Xinyu Zhang, Baowei Xv, Jinzhao Zhang, Rong Fu, Xiaoyu Wang, Lei Zhu, Haibing Ren, Pingping Lu, Jun Li, et~al.
\newblock Interfusion: Interaction-based 4d radar and lidar fusion for 3d object detection.
\newblock In {\em Proceedings of the IEEE/RSJ International Conference on Intelligent Robots and Systems (IROS)}, pages 12247--12253. IEEE, 2022.

\bibitem{wang2022multi}
Li~Wang, Xinyu Zhang, Jun Li, Baowei Xv, Rong Fu, Haifeng Chen, Lei Yang, Dafeng Jin, and Lijun Zhao.
\newblock Multi-modal and multi-scale fusion 3d object detection of 4d radar and lidar for autonomous driving.
\newblock {\em IEEE Transactions on Vehicular Technology}, 2022.

\bibitem{paek2023enhanced}
Dong-Hee Paek, Seung-Hyun Kong, and Kevin~Tirta Wijaya.
\newblock Enhanced k-radar: Optimal density reduction to improve detection performance and accessibility of 4d radar tensor-based object detection.
\newblock In {\em Proceedings of the IEEE Intelligent Vehicles Symposium (IV)}, pages 1--6. IEEE, 2023.

\bibitem{zheng2022tj4dradset}
Lianqing Zheng, Zhixiong Ma, Xichan Zhu, Bin Tan, Sen Li, Kai Long, Weiqi Sun, Sihan Chen, Lu~Zhang, Mengyue Wan, et~al.
\newblock Tj4dradset: A 4d radar dataset for autonomous driving.
\newblock In {\em Proceedings of the IEEE 25th International Conference on Intelligent Transportation Systems (ITSC)}, pages 493--498. IEEE, 2022.

\bibitem{liu2023smurf}
Jianan Liu, Qiuchi Zhao, Weiyi Xiong, Tao Huang, Qing-Long Han, and Bing Zhu.
\newblock Smurf: Spatial multi-representation fusion for 3d object detection with 4d imaging radar.
\newblock {\em IEEE Transactions on Intelligent Vehicles}, 2023.

\bibitem{zheng2023rcfusion}
Lianqing Zheng, Sen Li, Bin Tan, Long Yang, Sihan Chen, Libo Huang, Jie Bai, Xichan Zhu, and Zhixiong Ma.
\newblock Rcfusion: Fusing 4d radar and camera with bird’s-eye view features for 3d object detection.
\newblock {\em IEEE Transactions on Instrumentation and Measurement}, 2023.

\bibitem{xiong2023lxl}
Weiyi Xiong, Jianan Liu, Tao Huang, Qing-Long Han, Yuxuan Xia, and Bing Zhu.
\newblock Lxl: Lidar excluded lean 3d object detection with 4d imaging radar and camera fusion.
\newblock {\em IEEE Transactions on Intelligent Vehicles}, 2023.

\bibitem{zhang2024tl}
Haoyi Zhang, Kai Wu, Rongkang Chen, Zihao Wu, Yong Zhong, and Weihua Li.
\newblock Tl-4drcf: A two-level 4d radar-camera fusion method for object detection in adverse weather.
\newblock {\em IEEE Sensors Journal}, 2024.

\bibitem{kong2023rtnh+}
Seung-Hyun Kong, Dong-Hee Paek, and Sangjae Cho.
\newblock Rtnh+: Enhanced 4d radar object detection network using combined cfar-based two-level preprocessing and vertical encoding.
\newblock {\em arXiv preprint arXiv:2310.17659}, 2023.

\bibitem{yan2023mvfan}
Qiao Yan and Yihan Wang.
\newblock Mvfan: Multi-view feature assisted network for 4d radar object detection.
\newblock In {\em Proceedings of the International Conference on Neural Information Processing}, pages 493--511. Springer, 2023.

\bibitem{zhang2023dual}
Xinyu Zhang, Li~Wang, Jian Chen, Cheng Fang, Lei Yang, Ziying Song, Guangqi Yang, Yichen Wang, Xiaofei Zhang, and Jun Li.
\newblock Dual radar: A multi-modal dataset with dual 4d radar for autononous driving.
\newblock {\em arXiv preprint arXiv:2310.07602}, 2023.

\bibitem{deng2023see}
Jianning Deng, Gabriel Chan, Hantao Zhong, and Chris~Xiaoxuan Lu.
\newblock See beyond seeing: Robust 3d object detection from point clouds via cross-modal hallucination.
\newblock {\em arXiv preprint arXiv:2309.17336}, 2023.

\bibitem{cui20213d}
Hang Cui, Junzhe Wu, Jiaming Zhang, Girish Chowdhary, and William~R Norris.
\newblock 3d detection and tracking for on-road vehicles with a monovision camera and dual low-cost 4d mmwave radars.
\newblock In {\em Proceedings of the IEEE International Intelligent Transportation Systems Conference (ITSC)}, pages 2931--2937. IEEE, 2021.

\bibitem{pan2023moving}
Zhijun Pan, Fangqiang Ding, Hantao Zhong, and Chris~Xiaoxuan Lu.
\newblock Moving object detection and tracking with 4d radar point cloud.
\newblock In {\em Proceedings of the IEEE International Conference on Robotics and Automation (ICRA)}, 2024.

\bibitem{tan2023tracking}
Bin Tan, Zhixiong Ma, Xichan Zhu, Sen Li, Lianqing Zheng, Libo Huang, and Jie Bai.
\newblock Tracking of multiple static and dynamic targets for 4d automotive millimeter-wave radar point cloud in urban environments.
\newblock {\em Remote Sensing}, 15(11):2923, 2023.

\bibitem{scharf1991statistical}
Louis~L Scharf and C{\'e}dric Demeure.
\newblock {\em Statistical signal processing: detection, estimation, and time series analysis}.
\newblock Prentice Hall, 1991.

\bibitem{gandhi1988analysis}
Prashant~P Gandhi and Saleem~A Kassam.
\newblock Analysis of cfar processors in nonhomogeneous background.
\newblock {\em IEEE Transactions on Aerospace and Electronic systems}, 24(4):427--445, 1988.

\bibitem{zhu2020deformable}
Xizhou Zhu, Weijie Su, Lewei Lu, Bin Li, Xiaogang Wang, and Jifeng Dai.
\newblock Deformable detr: Deformable transformers for end-to-end object detection.
\newblock In {\em Proceedings of the International Conference on Learning Representations (ICLR)}, 2020.

\bibitem{roldao20223d}
Luis Roldao, Raoul De~Charette, and Anne Verroust-Blondet.
\newblock 3d semantic scene completion: A survey.
\newblock {\em International Journal of Computer Vision}, 130(8):1978--2005, 2022.

\bibitem{song2017semantic}
Shuran Song, Fisher Yu, Andy Zeng, Angel~X Chang, Manolis Savva, and Thomas Funkhouser.
\newblock Semantic scene completion from a single depth image.
\newblock In {\em Proceedings of the IEEE Conference on Computer Vision and Pattern Recognition (CVPR)}, pages 1746--1754, 2017.

\bibitem{liu2018see}
Shice Liu, Yu~Hu, Yiming Zeng, Qiankun Tang, Beibei Jin, Yinhe Han, and Xiaowei Li.
\newblock See and think: Disentangling semantic scene completion.
\newblock {\em Advances in Neural Information Processing Systems}, 31, 2018.

\bibitem{zhang2018efficient}
Jiahui Zhang, Hao Zhao, Anbang Yao, Yurong Chen, Li~Zhang, and Hongen Liao.
\newblock Efficient semantic scene completion network with spatial group convolution.
\newblock In {\em Proceedings of the European Conference on Computer Vision (ECCV)}, pages 733--749, 2018.

\bibitem{li2019depth}
Jie Li, Yu~Liu, Xia Yuan, Chunxia Zhao, Roland Siegwart, Ian Reid, and Cesar Cadena.
\newblock Depth based semantic scene completion with position importance aware loss.
\newblock {\em IEEE Robotics and Automation Letters}, 5(1):219--226, 2019.

\bibitem{li2019rgbd}
Jie Li, Yu~Liu, Dong Gong, Qinfeng Shi, Xia Yuan, Chunxia Zhao, and Ian Reid.
\newblock Rgbd based dimensional decomposition residual network for 3d semantic scene completion.
\newblock In {\em Proceedings of the IEEE/CVF Conference on Computer Vision and Pattern Recognition (CVPR)}, pages 7693--7702, 2019.

\bibitem{zhang2019cascaded}
Pingping Zhang, Wei Liu, Yinjie Lei, Huchuan Lu, and Xiaoyun Yang.
\newblock Cascaded context pyramid for full-resolution 3d semantic scene completion.
\newblock In {\em Proceedings of the IEEE/CVF International Conference on Computer Vision (ICCV)}, pages 7801--7810, 2019.

\bibitem{li2020anisotropic}
Jie Li, Kai Han, Peng Wang, Yu~Liu, and Xia Yuan.
\newblock Anisotropic convolutional networks for 3d semantic scene completion.
\newblock In {\em Proceedings of the IEEE/CVF Conference on Computer Vision and Pattern Recognition (CVPR)}, pages 3351--3359, 2020.

\bibitem{chen20203d}
Xiaokang Chen, Kwan-Yee Lin, Chen Qian, Gang Zeng, and Hongsheng Li.
\newblock 3d sketch-aware semantic scene completion via semi-supervised structure prior.
\newblock In {\em Proceedings of the IEEE/CVF Conference on Computer Vision and Pattern Recognition (CVPR)}, pages 4193--4202, 2020.

\bibitem{cai2021semantic}
Yingjie Cai, Xuesong Chen, Chao Zhang, Kwan-Yee Lin, Xiaogang Wang, and Hongsheng Li.
\newblock Semantic scene completion via integrating instances and scene in-the-loop.
\newblock In {\em Proceedings of the IEEE/CVF Conference on Computer Vision and Pattern Recognition (CVPR)}, pages 324--333, 2021.

\bibitem{behley2019semantickitti}
Jens Behley, Martin Garbade, Andres Milioto, Jan Quenzel, Sven Behnke, Cyrill Stachniss, and Jurgen Gall.
\newblock Semantickitti: A dataset for semantic scene understanding of lidar sequences.
\newblock In {\em Proceedings of the IEEE/CVF International Conference on Computer Vision (ICCV)}, pages 9297--9307, 2019.

\bibitem{ding2023hidden}
Fangqiang Ding, Andras Palffy, Dariu~M Gavrila, and Chris~Xiaoxuan Lu.
\newblock Hidden gems: 4d radar scene flow learning using cross-modal supervision.
\newblock In {\em Proceedings of the IEEE/CVF Conference on Computer Vision and Pattern Recognition}, pages 9340--9349, 2023.

\bibitem{ding2022self}
Fangqiang Ding, Zhijun Pan, Yimin Deng, Jianning Deng, and Chris~Xiaoxuan Lu.
\newblock Self-supervised scene flow estimation with 4-d automotive radar.
\newblock {\em IEEE Robotics and Automation Letters}, 7(3):8233--8240, 2022.

\bibitem{Ding_2024_ECCV}
Fangqiang Ding, Zhen Luo, Peijun Zhao, and Chris~Xiaoxuan Lu.
\newblock milliflow: Scene flow estimation on mmwave radar point cloud for human motion sensing.
\newblock In {\em Proceedings of the European Conference on Computer Vision (ECCV)}, 2024.

\bibitem{zhang2023ntu4dradlm}
Jun Zhang, Huayang Zhuge, Yiyao Liu, Guohao Peng, Zhenyu Wu, Haoyuan Zhang, Qiyang Lyu, Heshan Li, Chunyang Zhao, Dogan Kircali, et~al.
\newblock Ntu4dradlm: 4d radar-centric multi-modal dataset for localization and mapping.
\newblock In {\em Proceedings of the IEEE 26th International Conference on Intelligent Transportation Systems (ITSC)}, pages 4291--4296. IEEE, 2023.

\bibitem{choi2023msc}
Minseong Choi, Seunghoon Yang, Seungho Han, Yeongseok Lee, Minyoung Lee, Keun~Ha Choi, and Kyung-Soo Kim.
\newblock Msc-rad4r: Ros-based automotive dataset with 4d radar.
\newblock {\em IEEE Robotics and Automation Letters}, 2023.

\bibitem{lu2023efficient}
Shouyi Lu, Guirong Zhuo, Lu~Xiong, Xichan Zhu, Lianqing Zheng, Zihang He, Mingyu Zhou, Xinfei Lu, and Jie Bai.
\newblock Efficient deep-learning 4d automotive radar odometry method.
\newblock {\em IEEE Transactions on Intelligent Vehicles}, 2023.

\bibitem{zhuoins20234drvo}
Guirong Zhuoins, Shouyi Lu, Lu~Xiong, Huanyu Zhouins, Lianqing Zheng, and Mingyu Zhou.
\newblock 4drvo-net: Deep 4d radar--visual odometry using multi-modal and multi-scale adaptive fusion.
\newblock {\em IEEE Transactions on Intelligent Vehicles}, 2023.

\bibitem{zhuang20234d}
Yuan Zhuang, Binliang Wang, Jianzhu Huai, and Miao Li.
\newblock 4d iriom: 4d imaging radar inertial odometry and mapping.
\newblock {\em IEEE Robotics and Automation Letters}, 2023.

\bibitem{li20234d}
Xingyi Li, Han Zhang, and Weidong Chen.
\newblock 4d radar-based pose graph slam with ego-velocity pre-integration factor.
\newblock {\em IEEE Robotics and Automation Letters}, 2023.

\bibitem{zhang20234dradarslam}
Jun Zhang, Huayang Zhuge, Zhenyu Wu, Guohao Peng, Mingxing Wen, Yiyao Liu, and Danwei Wang.
\newblock 4dradarslam: A 4d imaging radar slam system for large-scale environments based on pose graph optimization.
\newblock In {\em Proceedings of the IEEE International Conference on Robotics and Automation (ICRA)}, pages 8333--8340. IEEE, 2023.

\bibitem{rebut2022raw}
Julien Rebut, Arthur Ouaknine, Waqas Malik, and Patrick P{\'e}rez.
\newblock Raw high-definition radar for multi-task learning.
\newblock In {\em Proceedings of the IEEE/CVF Conference on Computer Vision and Pattern Recognition (CVPR)}, pages 17021--17030, 2022.

\bibitem{zhang2020object}
Guoqiang Zhang, Haopeng Li, and Fabian Wenger.
\newblock Object detection and 3d estimation via an fmcw radar using a fully convolutional network.
\newblock In {\em Proceedings of the IEEE International Conference on Acoustics, Speech and Signal Processing (ICASSP)}, pages 4487--4491. IEEE, 2020.

\bibitem{wang2021rodnet}
Yizhou Wang, Zhongyu Jiang, Yudong Li, Jenq-Neng Hwang, Guanbin Xing, and Hui Liu.
\newblock Rodnet: A real-time radar object detection network cross-supervised by camera-radar fused object 3d localization.
\newblock {\em IEEE Journal of Selected Topics in Signal Processing}, 15(4):954--967, 2021.

\bibitem{dong2020probabilistic}
Xu~Dong, Pengluo Wang, Pengyue Zhang, and Langechuan Liu.
\newblock Probabilistic oriented object detection in automotive radar.
\newblock In {\em Proceedings of the IEEE/CVF Conference on Computer Vision and Pattern Recognition Workshops (CVPRW)}, pages 102--103, 2020.

\bibitem{zhang2021raddet}
Ao~Zhang, Farzan~Erlik Nowruzi, and Robert Laganiere.
\newblock Raddet: Range-azimuth-doppler based radar object detection for dynamic road users.
\newblock In {\em Proceedings of the 18th Conference on Robots and Vision (CRV)}, pages 95--102. IEEE, 2021.

\bibitem{palffy2020cnn}
Andras Palffy, Jiaao Dong, Julian~FP Kooij, and Dariu~M Gavrila.
\newblock Cnn based road user detection using the 3d radar cube.
\newblock {\em IEEE Robotics and Automation Letters}, 5(2):1263--1270, 2020.

\bibitem{major2019vehicle}
Bence Major, Daniel Fontijne, Amin Ansari, Ravi Teja~Sukhavasi, Radhika Gowaikar, Michael Hamilton, Sean Lee, Slawomir Grzechnik, and Sundar Subramanian.
\newblock Vehicle detection with automotive radar using deep learning on range-azimuth-doppler tensors.
\newblock In {\em Proceedings of the IEEE/CVF International Conference on Computer Vision Workshops (ICCVW)}, 2019.

\bibitem{TIMmWaveRadar2024}
{Texas Instruments}.
\newblock {mmWave Radar Sensors - Overview}.
\newblock \url{https://www.ti.com/sensors/mmwave-radar/overview.html}, 2024.
\newblock Accessed: 2024-02-22.

\bibitem{kramer2022coloradar}
Andrew Kramer, Kyle Harlow, Christopher Williams, and Christoffer Heckman.
\newblock Coloradar: The direct 3d millimeter wave radar dataset.
\newblock {\em The International Journal of Robotics Research}, 41(4):351--360, 2022.

\bibitem{blake1988cfar}
Stephen Blake.
\newblock Os-cfar theory for multiple targets and nonuniform clutter.
\newblock {\em IEEE Transactions on Aerospace and Electronic Systems}, 24(6):785--790, 1988.

\bibitem{cheng2022novel}
Yuwei Cheng, Jingran Su, Mengxin Jiang, and Yimin Liu.
\newblock A novel radar point cloud generation method for robot environment perception.
\newblock {\em IEEE Transactions on Robotics}, 38(6):3754--3773, 2022.

\bibitem{tran2015learning}
Du~Tran, Lubomir Bourdev, Rob Fergus, Lorenzo Torresani, and Manohar Paluri.
\newblock Learning spatiotemporal features with 3d convolutional networks.
\newblock In {\em Proceedings of the IEEE International Conference on Computer Vision (ICCV)}, pages 4489--4497, 2015.

\bibitem{zhou2018voxelnet}
Yin Zhou and Oncel Tuzel.
\newblock Voxelnet: End-to-end learning for point cloud based 3d object detection.
\newblock In {\em Proceedings of the IEEE Conference on Computer Vision and Pattern Recognition (CVPR)}, pages 4490--4499, 2018.

\bibitem{kwok2015effects}
R~Kwok and C~Haas.
\newblock Effects of radar side-lobes on snow depth retrievals from operation icebridge.
\newblock {\em Journal of Glaciology}, 61(227):576--584, 2015.

\bibitem{tait2005introduction}
Peter Tait.
\newblock {\em Introduction to radar target recognition}, volume~18.
\newblock IET, 2005.

\bibitem{nie2023partner}
Ming Nie, Yujing Xue, Chunwei Wang, Chaoqiang Ye, Hang Xu, Xinge Zhu, Qingqiu Huang, Michael~Bi Mi, Xinchao Wang, and Li~Zhang.
\newblock Partner: Level up the polar representation for lidar 3d object detection.
\newblock In {\em Proceedings of the IEEE/CVF International Conference on Computer Vision (ICCV)}, pages 3801--3813, 2023.

\bibitem{zhu2021cylindrical}
Xinge Zhu, Hui Zhou, Tai Wang, Fangzhou Hong, Yuexin Ma, Wei Li, Hongsheng Li, and Dahua Lin.
\newblock Cylindrical and asymmetrical 3d convolution networks for lidar segmentation.
\newblock In {\em Proceedings of the IEEE/CVF Conference on Computer Vision and Pattern Recognition (CVPR)}, pages 9939--9948, 2021.

\bibitem{vaswani2017attention}
Ashish Vaswani, Noam Shazeer, Niki Parmar, Jakob Uszkoreit, Llion Jones, Aidan~N Gomez, {\L}ukasz Kaiser, and Illia Polosukhin.
\newblock Attention is all you need.
\newblock {\em Advances in neural information processing systems}, 30, 2017.

\bibitem{yan2018second}
Yan Yan, Yuxing Mao, and Bo~Li.
\newblock Second: Sparsely embedded convolutional detection.
\newblock {\em Sensors}, 18(10):3337, 2018.

\bibitem{he2016deep}
Kaiming He, Xiangyu Zhang, Shaoqing Ren, and Jian Sun.
\newblock Deep residual learning for image recognition.
\newblock In {\em Proceedings of the IEEE Conference on Computer Vision and Pattern Recognition (CVPR)}, pages 770--778, 2016.

\bibitem{lin2017feature}
Tsung-Yi Lin, Piotr Doll{\'a}r, Ross Girshick, Kaiming He, Bharath Hariharan, and Serge Belongie.
\newblock Feature pyramid networks for object detection.
\newblock In {\em Proceedings of the IEEE Conference on Computer Vision and Pattern Recognition (CVPR)}, pages 2117--2125, 2017.

\bibitem{berman2018lovasz}
Maxim Berman, Amal~Rannen Triki, and Matthew~B Blaschko.
\newblock The lov{\'a}sz-softmax loss: A tractable surrogate for the optimization of the intersection-over-union measure in neural networks.
\newblock In {\em Proceedings of the IEEE Conference on Computer Vision and Pattern Recognition (CVPR)}, pages 4413--4421, 2018.

\bibitem{li2023sscbench}
Yiming Li, Sihang Li, Xinhao Liu, Moonjun Gong, Kenan Li, Nuo Chen, Zijun Wang, Zhiheng Li, Tao Jiang, Fisher Yu, et~al.
\newblock Sscbench: A large-scale 3d semantic scene completion benchmark for autonomous driving.
\newblock {\em arXiv preprint arXiv:2306.09001}, 2023.

\bibitem{spconv2022}
{Spconv Contributors}.
\newblock Spconv: Spatially sparse convolution library.
\newblock \url{https://github.com/traveller59/spconv}, 2022.

\bibitem{Ilya2017Stochastic}
Ilya Loshchilov and Frank Hutter.
\newblock {SGDR:} stochastic gradient descent with warm restarts.
\newblock In {\em Proceedings of the International Conference on Learning Representations (ICLR)}, 2017.

\end{thebibliography}

% References follow the acknowledgments in the camera-ready paper. Use unnumbered first-level heading for
% the references. Any choice of citation style is acceptable as long as you are
% consistent. It is permissible to reduce the font size to \verb+small+ (9 point)
% when listing the references.
% Note that the Reference section does not count towards the page limit.
% \medskip

% {
% \small

% [1] Alexander, J.A.\ \& Mozer, M.C.\ (1995) Template-based algorithms for
% connectionist rule extraction. In G.\ Tesauro, D.S.\ Touretzky and T.K.\ Leen
% (eds.), {\it Advances in Neural Information Processing Systems 7},
% pp.\ 609--616. Cambridge, MA: MIT Press.

% [2] Bower, J.M.\ \& Beeman, D.\ (1995) {\it The Book of GENESIS: Exploring
%   Realistic Neural Models with the GEneral NEural SImulation System.}  New York:
% TELOS/Springer--Verlag.

% [3] Hasselmo, M.E., Schnell, E.\ \& Barkai, E.\ (1995) Dynamics of learning and
% recall at excitatory recurrent synapses and cholinergic modulation in rat
% hippocampal region CA3. {\it Journal of Neuroscience} {\bf 15}(7):5249-5262.
% }

%%%%%%%%%%%%%%%%%%%%%%%%%%%%%%%%%%%%%%%%%%%%%%%%%%%%%%%%%%%%
\newpage
\appendix

\section*{Appendix}
The appendix is organized as follows:
\begin{itemize}[label=$\bullet$]
    \setlength{\itemsep}{0pt}
    \setlength{\parsep}{0pt}
    \setlength{\parskip}{0pt}
    \item Section~\ref{sec:setup} illustrates more details on our experiment setup, including ground truth generation, dataset statistics, evaluation area and computation resources we used for our experiments.
    \item Section~\ref{sec:details} introduces implementation details of different components in {\sysname}.
    \item Section~\ref{sec:visualization} gives more 
    experimental results, visualization and failure case of {\sysname}.
\end{itemize}
Besides, please refer to our supplementary video for more qualitative results.

\section{Experiment setup details}\label{sec:setup}

\noindent\textbf{Ground truth generation.}
Our pipeline of 3D occupancy annotation is similar to those in~\cite{wang2023openoccupancy,wei2023surroundocc,li2023sscbench,tong2023scene}. First, we split each LiDAR point cloud from a sequence into the background and foreground part with the help of 3D bounding box annotations. For the background, we superimpose all LiDAR points by transforming them into a united world coordinate using their extrinsic. For the foreground part, we track the same instances (indicated by the same tracking IDs) across the sequence and transform LiDAR points association to them into the coordinates of their bounding boxes. In this way, sparse LiDAR point clouds can be significantly densified and the occupancy labels can be more realistic. Note that K-Radar~\cite{paek2022k} only annotates the objects in the front of the car. To avoid the interference of moving objects in the back, we only use the front part of each LiDAR sweep for ground truth generation. Second, we transform the background and objects point sets into the current frame coordinate system with respect to the ego-pose of the current frame and the objects' pose. Lastly, we concatenate the background and objects points at the current frame and voxelized the merged point cloud to generate the occupancy labels. In cases where voxels are overlaid or boundaries are not clear, we use the majority voting to decide voxel-wise semantics (foreground vs. background). 

\noindent\textbf{Dataset statistics.} In adverse weather conditions (\emph{e.g.,} fog, rain and snow), water droplets or snowflakes can scatter or absorb LiDAR beams, reducing the effective range of LiDAR and inducing noise in the data. To ensure the high fidelity of our occupancy labels, we select 24 sequences collected in decent weather conditions from K-Radar~\cite{paek2022k} for annotation and leave the rest sequences collected in poor weathers unannotated, which can only be used for qualitative analysis. We split the annotated 24 sequences into the training, validation and test sets with a ratio of 17:2:5, resulting in 11,333, 1,059 and 2,878 frames, respective. Over
0.5 billion voxels are obtained from all annotated frames, among which free, background and foreground class accounts for 92.3\%, 7.4\% and 0.3\% individually.

\noindent\textbf{Evaluation area.} As claimed in the main paper, we only report the evaluation results within the area where the horizontal FoV (hFov) of all sensors overlap. This scheme can lead to a more fair comparison as it avoids assessing the hallucinated voxels beyond hFoV for modalities like radar and camera, whose hFoVs cannot fully cover our defined RoI volume ahead of the car. Specifically, the overleap hFoV of K-Radar~\cite{paek2022k} sensor suite is $107^{\circ}$, symmetrically distributed around the front axis. The ratio  between the final evaluation area and our RoI is calculated as: $1 - \cot\left({107^\circ}/{2}\right)/{4} \approx 0.812$.

\noindent\textbf{Computation resources.} All of our experiments are conducted on a Ubuntu server equipped with 2 Nvidia RTX 3090 - 24GB  GPUs, an Intel i9-10980XE CPU @ 3.00GHz and a 64GB RAM. The training of our method {\sysname}  uses 17.98GB VRAM, and takes approximately 16.7 hours.

\noindent\textbf{License for K-Radar.} The K-Radar dataset~\cite{paek2022k} is published under the CC BY-NC-ND License, and all K-Radar codes~\footnote{https://github.com/kaist-avelab/K-Radar} are published under the Apache License 2.0.
\section{Implementation details of RadarOcc}\label{sec:details}

\noindent\textbf{Data volume reduction.} The volume size of input raw 4DRT $\mathbf{V}$ is $256\times 107\times 37\times 64~(R\times A\times E\times D)$. By encoding the Doppler bins for each spatial location into 8-channel descriptors, we reduce the size of 4DRTs by $\times\frac{D}{8}$, leading to a 3D spatial data volume with a size of $(256\times 107\times 37)\times 8$ with the Doppler axis as the feature dimension. For sidelobe-aware spatial sparsifying, we select the top-$N_r$ ($N_r=250$) elements per range. The resulting lightweight sparse RT $\mathbf{T}$ per frame is $\sim$5MB. Please refer to Sec.~\ref{elements} for how we select the optimal $N_r$.

\noindent\textbf{Range-wise self-attention.} In our spherical-based feature encoding, the range-wise self-attention is performed on the non-empty elements per range, \emph{i.e.,} 
$t_i\in\mathbb{R}^{N_r \times (8+2)} (i=1,2,\dots,R)$, where $N_r = 250$. The 8-channel Doppler descriptors are considered as the input features while the azimuth and elevation indices are converted to positional embeddings with lookup tables~\cite{vaswani2017attention}. Specifically, we use two layers of multi-head attention with the embedding dimension set as 32, number of heads as 4 and dropout probability to be 0.1 The output is re-organized to a sparse tensor with a dimension of $RN_r\times (32+3)$, where the range, azimuth and elevation index is stored for each non-empty element.
% For each token, the feature is projected to key, query and value via linear projection with a hidden dimension of $C_{hidden} = 32$ and Dot-Product is used for the attention mechanism. %To Modify

\noindent\textbf{3D sparse convolution.}
We utilize the \texttt{spconv} library~\cite{spconv2022} to implement the sparse convolution layers for our spherical-based feature encoding. This encoding process has two types of operation: \emph{3D Submanifold Convolution} and \emph{3D Sparse Convolution}. 3D submanifold convolution only convolves the active spatial locations without altering the sparsity pattern and spatial resolution, while 3D sparse convolution performs convolution on all active locations, expanding the sparsify pattern and allows for spatial resolution change. We leverage the submanifold convolution as the first and last layer and apply three sparse convolution layers in-between. We set the stride as 2 for the last two of 3D sparse convolution to reduce the spatial dimension. As a result, we obtain a 3D dense feature volume $\mathbf{F}\in\mathbb{R}^{64\times 27\times 10\times C_f} (C_f = 192)$, where the spatial dimension is decreased by $\times 4$.

\noindent\textbf{Deformable self-attention.}  Given feature volume $\mathbf{F}$, we use 3D deformable self-attention~\cite{zhu2020deformable} to augment its spatial features. 
 Two attention layers are used and the number of sampling points of each query is set to 8. Each attention layer has 8 heads and apply a dropout of a rate of 0.1 to the output features. The refined feature volume $\mathbf{F}_r$ has the same dimension as $\mathbf{F}$, \emph{i.e.,} $64\times 27\times 10\times C_f$.
 
\noindent\textbf{Spherical-to-Cartesian feature aggregation.} To aggregate features extracted in the spherical coordinates, we build learnable voxel queries $\mathbf{H}=\{h^q\}_q$ with a dimension of ${14\times 128\times 128\times C_f}$ defined in the LiDAR Cartesian coordinates system. By transforming the 3D points $p^q$ corresponding to our voxel queries $h^q$ into the radar spherical coordinates, we construct a list of 3D reference points $\Phi (p^q)$. Then, the deformable cross attention is used to aggregate the spherical features to Cartesian by considering the spherical volume $\mathbf{F}_r$ as the key and value of attention and the voxel queries $\mathbf{H}$ as the query of the attention. Just as the self-attention module, we use the 3D version of the deformable attention~\cite{zhu2020deformable}, with the same network settings. The dimension of the output Cartesian feature $\mathbf{G}$ have the same size as the learnable queries  $\mathbf{H}$, which is $14\times 128\times 128\times C_f$.

\noindent\textbf{3D occupancy decoding.} Given the Cartesian voxel features $\mathbf{G}$, we firstly apply the 3D version of ResNet-18~\cite{he2016deep} for decoding, resulting in 4 level of feature maps, with $ \frac{1}{2},\frac{1}{4},\frac{1}{8},\frac{1}{16}$ of the voxel spatial shape and  $80, 160, 320, 640$ for feature dimension respectively. These multi-level features are then upsampled back to the target spatial space $H\times W\times L$ using 3D FPN~\cite{lin2017feature}, leading to the final features $\mathbf{G}_d$ with a dimension of $14\times 128\times 128\times 4C_f$.  
Lastly, we use an MLP with the hidden dimension of [64,64] to reduce the feature channel and predict the occupancy probabilities which are normalized by a \emph{softmax} layer. The output is denoted as $\tilde{\mathbf{O}}\in\{0,1\}^{H\times W\times L\times (C+1)}$.

\noindent\textbf{Training loss.}
The overall loss function $\mathcal{L}$ used to train our network can be written as:
\begin{equation}\label{eqn:loss}
    \mathcal{L} = \mathcal{L}_{CE} + \mathcal{L}_{LS} + \mathcal{L}_{scal}^{geo} + \mathcal{L}_{scal}^{sem}
\end{equation}
Given the ground truth denoted as $\hat{\mathbf{O}}=\{\hat{o}_i\in\{c_0, c_1, \dots, c_C\}\}_{i=1}^{N_o} (N_o=H\times W\times L)$ and the output $\tilde{\mathbf{O}}$, the cross-entropy loss $\mathcal{L}_{CE}$ can be calculated as:
\begin{equation}
    \mathcal{L}_{CE} = - \sum_{i=1}^{N_o}\sum_{c=c_0}^{c_C} w_c\hat{o}_{i,c}\mathrm{log}(\tilde{o}_{i,c})
    % \left(\frac{e^{\tilde{o}_{i,c}}}{\sum_{c}e^{\tilde{o}_{i,c}}}\right)
\end{equation}
where $N_o$ is the number of voxels, $c$ and $i$ indexes classes and voxels, $\tilde{o}_{i,c}$ is the predicted logit for $i$-th voxel on the class $c$. $\hat{o}_{i,c}=1$ if $\hat{o}_{i}=c$; else, $\hat{o}_{i,c}=0$. To balance different classes, we use $w_c$ for each class calculated as the inverse of the class frequency in K-Radar~\cite{paek2022k}. Please refer to~\cite{berman2018lovasz} and \cite{cao2022monoscene} for more details on the lovasz-softmax loss $\mathcal{L}_{LS}$ and scene-class affinity loss $\mathcal{L}_{scal}^{geo}$ and $\mathcal{L}_{scal}^{sem}$ we used in Eq.~\ref{eqn:loss}.

\noindent\textbf{Training details.} We train RadarOcc with 10 epochs using Adam optimizer with a learning rate of 3e-4. The batch size is 1 for each GPU. We follow \cite{wang2023openoccupancy} to use loss normalization to balance the weight of the 4 different losses, and cosine annealing \cite{Ilya2017Stochastic} with $\frac{1}{3}$ warm-up ratio is used at the start of the training. 
% is set to have stride $S = 2$ and double the channel. An input projection submanifold layer is used to project the feature  $\mathbf{F}\in\mathbb{R}^{R\times A\times E\times C_{base}}.$ where base channel $C_{base} = 16$ Then the backbone is designed as:
% $\{SparseConv,norm,relu,[submanifold,norm,relu]\times4 \}\times2$.
% It result in a dense spherical volume $\mathbf{F^`}\in\mathbb{R}^{\frac{R}{4}\times \frac{A}{4}\times \frac{E}{4}\times 4C_{base} }$. 
\section{Additional experiment results}\label{sec:visualization}

\begin{table*}[!tb]\small
    \renewcommand\arraystretch{1}
    \setlength\tabcolsep{5pt}
    \centering
    \resizebox{\textwidth}{!}{
    \begin{tabular}{l|ccc|ccc|ccc|ccc|c}
        \toprule
          & \multicolumn{3}{c|}{IoU (\%)} & \multicolumn{3}{c|}{mIoU (\%)} & \multicolumn{3}{c|}{\tikz \fill [bg] (0,0) rectangle (0.6em,0.6em); BG IoU (\%)} & \multicolumn{3}{c}{\tikz \fill [fg] (0,0) rectangle (0.6em,0.6em); FG IoU (\%)} & \multirow{2}{*}{fps} \\
        \cmidrule(lr){2-4} \cmidrule(lr){5-7} \cmidrule(lr){8-10} \cmidrule(lr){11-13}
        $N_r$ & 12.8m & 25.6m & 51.2m & 12.8m & 25.6m & 51.2m & 12.8m & 25.6m & 51.2m & 12.8m & 25.6m & 51.2m &  \\ 
        \midrule 
        850 & - & - & - & - & - & - & - & - & - & - & -& - & CUDA OOM\\
        650 & 52.5& 43.9 &30.6& 34.4& \textbf{27.2} &19.7& 52.1 &43.7& 30.4 &16.7& \textbf{10.7}& 8.9 & 2.9\\
        450 & 53.9 & 44.3 & 30.9 & \textbf{36.8} & 26.9 & \textbf{19.9} & {53.7} & 44.0 & 30.6 & \textbf{19.9} & 9.7 & \textbf{9.2} & 3.1\\
        250 & \textbf{54.1} & \textbf{45.1} & \textbf{31.9} & 34.0 & 25.7 & 19.1 & \textbf{53.7} & \textbf{44.8} & \textbf{31.6} & 14.2 & 6.7 & 6.6 & 3.3	\\
        50 & 52.7 & 44.5 & 31.9 & 32.6 & 25.8 & 19.4 & 52.6 & 44.3 & 31.5 & 12.5 & 7.3 & 7.3 & 3.6\\
        \bottomrule
    \end{tabular}
    }
    \caption{Impact of the number of selected top elements per range (\emph{i.e.}, $N_r$) in our sidelobe-aware spatial sparsifying. The results are reported on the validation set. Best result is shown in \textbf{bold}.}
    \label{tab:N_r}
\end{table*}

\begin{table*}[!tb]\small
    \renewcommand\arraystretch{1}
    \setlength\tabcolsep{5pt}
    \centering
    \resizebox{0.4\textwidth}{!}{
    \begin{tabular}{c|c|c}
        \toprule
            $N_d$ & IoU @ 51.2m (\%) & mIoU @ 51.2m (\%) \\
        \midrule
            1 & 30.9 & 18.7 \\
            2 & 28.8 & \textbf{19.4} \\
            3 & \textbf{31.9} & 19.1 \\
            4 & 31.1 & 18.9 \\
            5 & 30.1 & 18.8 \\
        \bottomrule
    \end{tabular}
    }
    \caption{Impact of the number of reserved Doppler bins for each spatial location (\emph{i.e.}, $N_d$). The results are reported on the validation set. Best result is shown in \textbf{bold}.}
    \label{tab:N_d}
\end{table*}

\subsection{Impact of the number of reserved top elements $N_r$}\label{elements}
In Sec.~\ref{reduction}, we propose a sidelobe-aware spatial sparsification technique that selects the top-$N_r$ elements for each individual range rather than the entire dense radar tensor (RT).
There is indeed a trade-off between preserving critical measurements and filtering noise/compressing the radar tensor in this process. Excessive compression/filtering may result in the loss of weak reflections, while insufficient compression/filtering increases computational costs and retains some level of noise. 

To identify the optimal balance, we conducted a series of experiments varying the number of selected top elements for each range, \emph{i.e.}, $N_r$, and assessed performance and inference speed on the validation set. The results, presented in Table~\ref{tab:N_r}, indicate that {\sysname} achieves the best results in half of all metrics on our validation set when 
$N_r$ = 250. Both higher and lower values of 
 lead to suboptimal results, suggesting that 
 $N_r$ = 250 strikes the best balance between retaining critical signals and filtering noise. Additionally, the inference speed at 
 $N_r$ = 250 is relatively higher compared to configurations with larger 
 values. Therefore, we select 
 $N_r$ = 250 for {\sysname}’s evaluation on our testing set.

\subsection{Impact of the number of reserved Doppler bins $N_d$}
To investigate the effect of the number of preserved top values (\emph{i.e.,} $N_d$) among Doppler bins for each spatial location, we conducted a series of experiments by varying $N_d$. As shown in Table~\ref{tab:N_d}, the change in $N_d$
 does not significantly impact our results. For both efficiency and performance, we chose 
$N_d = 3$ for our method based on the validation set performance.

This can be explained by the fact that K-Radar~\cite{paek2022k} wraps around overflow values in Doppler measurements due to the limited Doppler measurement range. For example, Doppler speeds of 3.0 m/s and 6.0 m/s are measured within the range of -1.92 to 1.92 m/s as 3.0 - 3.84 = -0.84m/s and 6.0 - 3.84$\times$2 = -1.68m/s, respectively. This ambiguity means the information from the Doppler axis only marginally improves our model. Consequently, changing 
 hardly affects our performance. Table.~\ref{tab:ablation} in our paper also shows that our baseline without Doppler bin descriptor (w/o DBD), which only uses mean power, reflects this minimal impact. However, we believe our Doppler bin encoding method could bring more improvement with other radar sensors that have a larger measurement range.

\subsection{Impact of range-wise self-attention}
To verify the effectiveness of the range-wise self-attention used in our sphercial-based feature encoding (\emph{c.f.} Sec.~\ref{spherical}), we experiment by removing it from the network and show the results in Tab. \ref{fig:ablation_appendix}. It can be seen that {\sysname} improves the performance on most metrics by adding the range-wise self-attention. In particular on FG IoU, the relative gain is 14.4\%, 14.7\% and 11.3\% for the 3D volume of 12.8m, 25.6m and 51.2m, respectively. We credit this to the ability of range-wise self-attention to further suppress the sidelobe noises appearing around the foreground objects.

\subsection{Qualitative results under adverse weather}
To better show the qualitative results of {\sysname} and baseline methods based on other modalities, we make some video demos under different weather conditions and submit them as a supplementary material. We recommend our audience to watch the video for a better understanding of our work.

\begin{table*}[!tb]\small
    \renewcommand\arraystretch{1}
    \setlength\tabcolsep{6pt}
    \centering
    \resizebox{\textwidth}{!}{
    \begin{tabular}{l|l|ccc|ccc|ccc|ccc}
    \toprule
        & & \multicolumn{3}{c|}{IoU (\%)} & \multicolumn{3}{c|}{mIoU (\%)} & \multicolumn{3}{c|}{\tikz \fill [bg] (0,0) rectangle (0.6em,0.6em); BG IoU (\%)} & \multicolumn{3}{c}{\tikz \fill [fg] (0,0) rectangle (0.6em,0.6em); FG IoU (\%)} \\
    \midrule
        & Method & 12.8m & 25.6m & 51.2m & 12.8m & 25.6m & 51.2m & 12.8m & 25.6m & 51.2m & 12.8m & 25.6m & 51.2m \\
    \midrule
        (a) & Ours & \textbf{48.8} & \textbf{39.1} & 30.4 & \textbf{34.3} & \textbf{28.5} & \textbf{22.6} & \textbf{47.9} & \textbf{38.2} & 29.4 & \textbf{20.7} & \textbf{18.7} & \textbf{15.8} \\
        (b) & Ours w/o RWA & 48.6 & 39.0 & \textbf{30.7} & 32.8 & 27.7 & 22.0 & 47.4 & 38.0 & \textbf{29.6} & 18.1 & 16.3 & 14.2 \\
    \bottomrule
    \end{tabular}
    }
    \caption{Ablation studies on range-wise self-attention designs of {\sysname}.}
    \label{fig:ablation_appendix}
    \vspace{-1em}
\end{table*}

\begin{figure}[!t]
    \centering
    \begin{subfigure}[b]{0.3\textwidth}
        \centering
        \includegraphics[width=\textwidth]{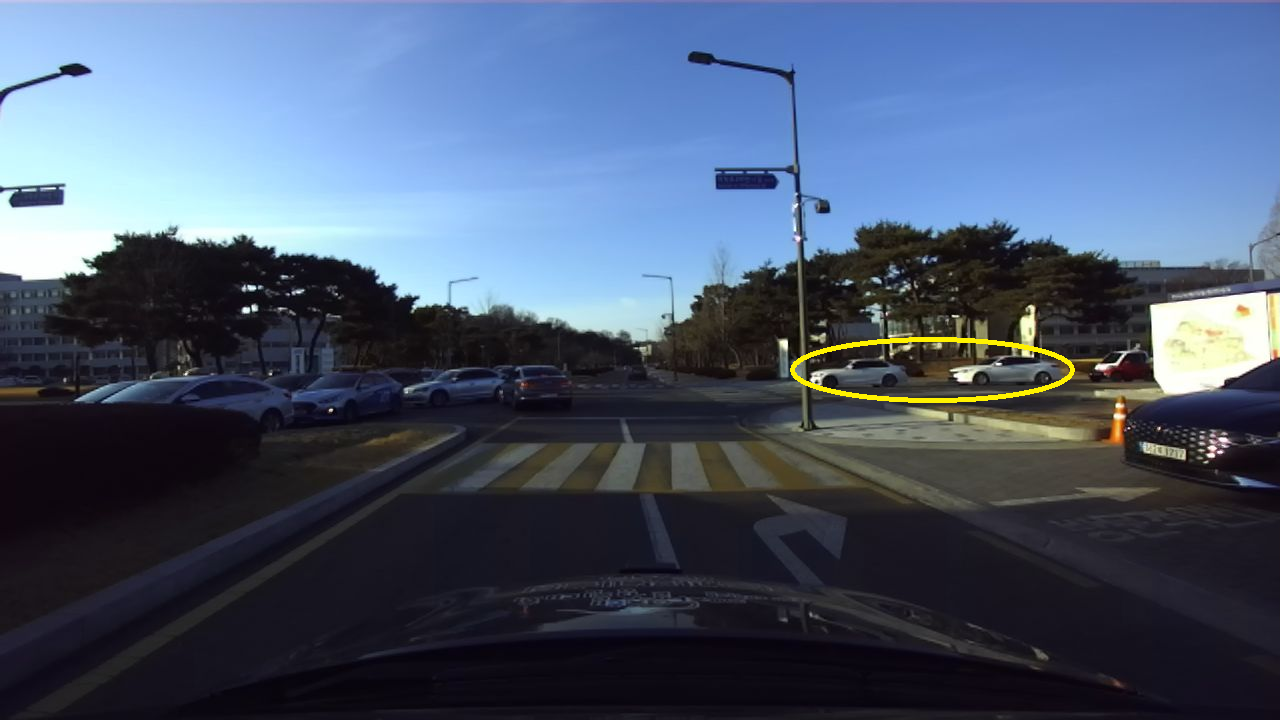}
        \caption{Camera View}
        \label{fig:image1}
    \end{subfigure}
    \hfill
    \begin{subfigure}[b]{0.3\textwidth}
        \centering
        \includegraphics[width=\textwidth]{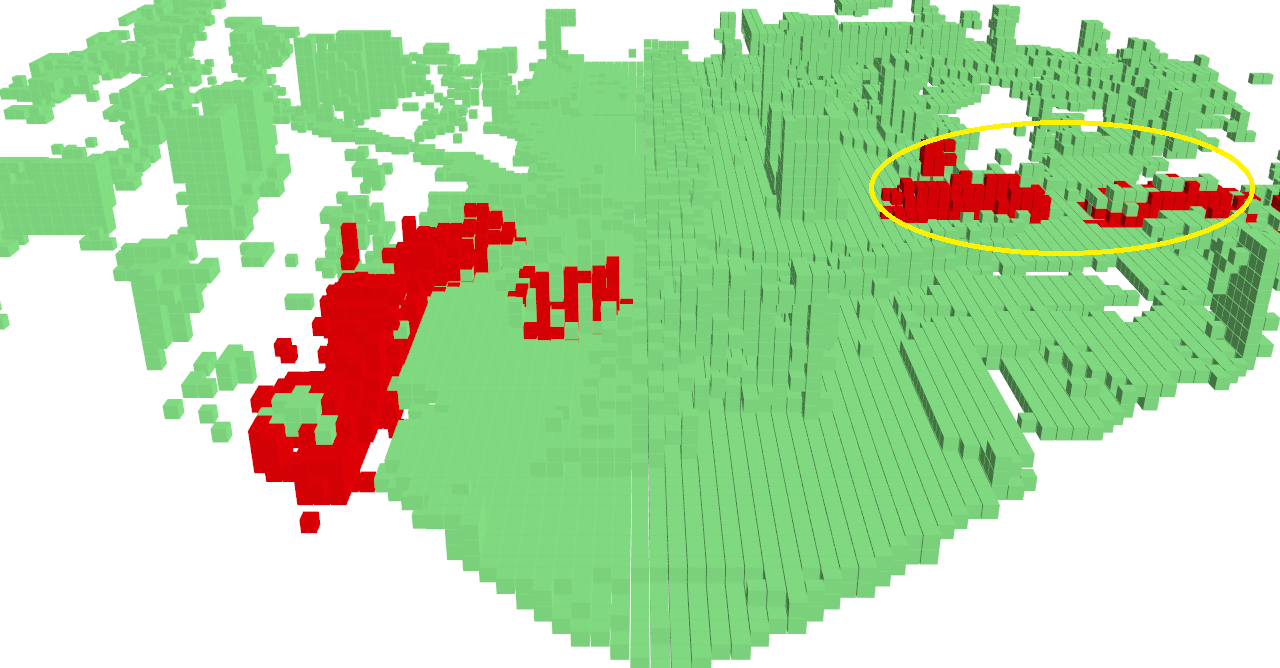}
        \caption{Ground truth}
        \label{fig:image2}
    \end{subfigure}
    \hfill
    \begin{subfigure}[b]{0.3\textwidth}
        \centering
        \includegraphics[width=\textwidth]{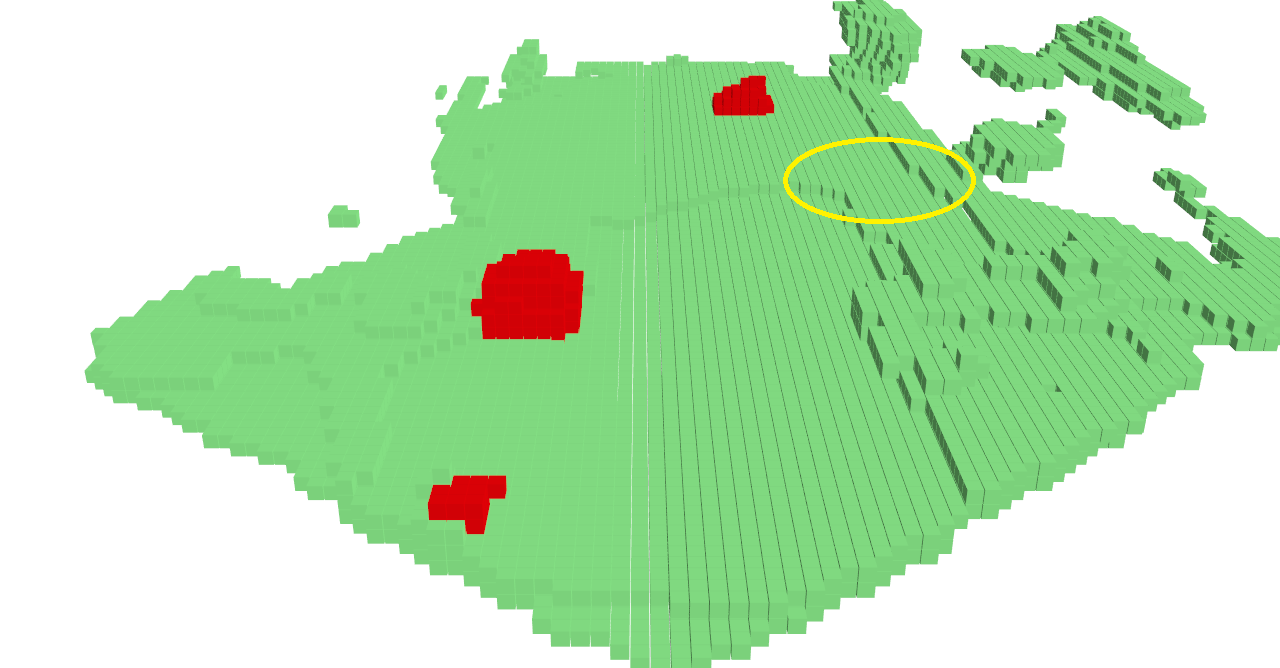}
        \caption{RadarOcc prediction}
        \label{fig:image3}
    \end{subfigure}
    \caption{Example of failure case due to insufficient resolution and decreased Signal-to-Noise Ratio at far distances. The white cars parked at the far right are not well predicted.}
    \label{fig:falure_1}
\end{figure}

\begin{figure}[!t]
    \centering
    \begin{subfigure}[b]{0.3\textwidth}
        \centering
        \includegraphics[width=\textwidth]{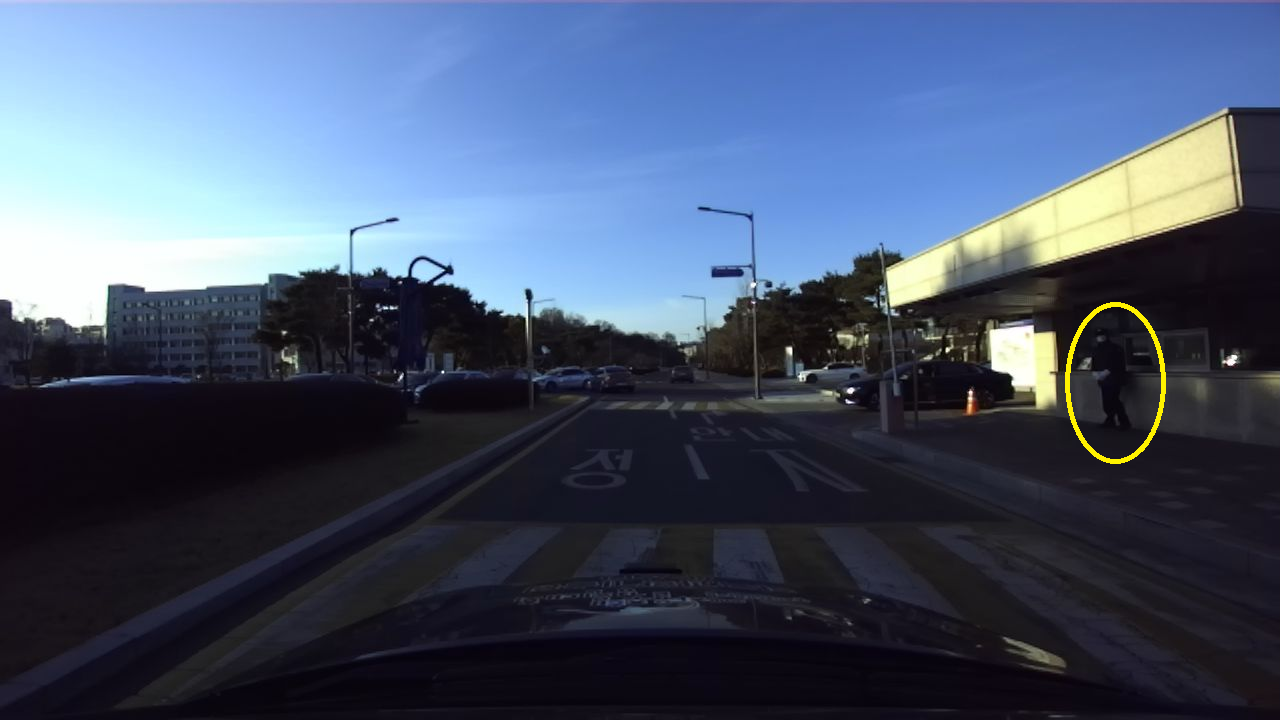}
        \caption{Camera View}
        \label{fig:image1}
    \end{subfigure}
    \hfill
    \begin{subfigure}[b]{0.3\textwidth}
        \centering
        \includegraphics[width=\textwidth]{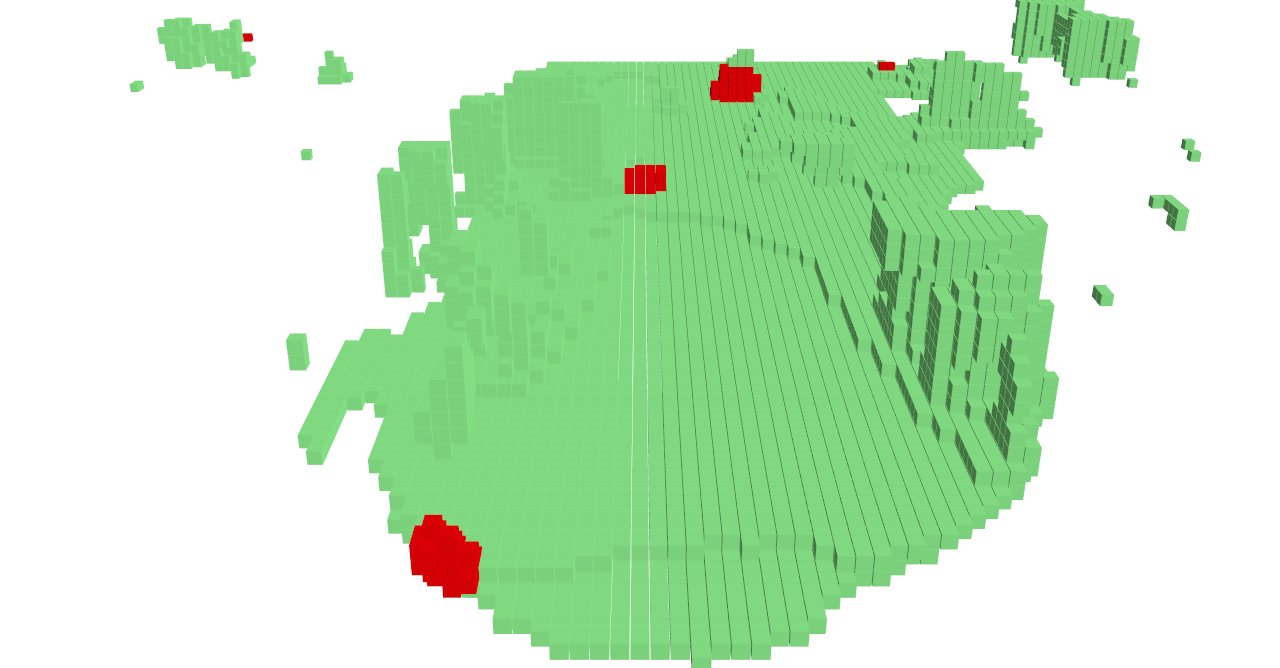}
        \caption{32 line LiDAR prediction}
        \label{fig:image2}
    \end{subfigure}
    \hfill
    \begin{subfigure}[b]{0.3\textwidth}
        \centering
        \includegraphics[width=\textwidth]{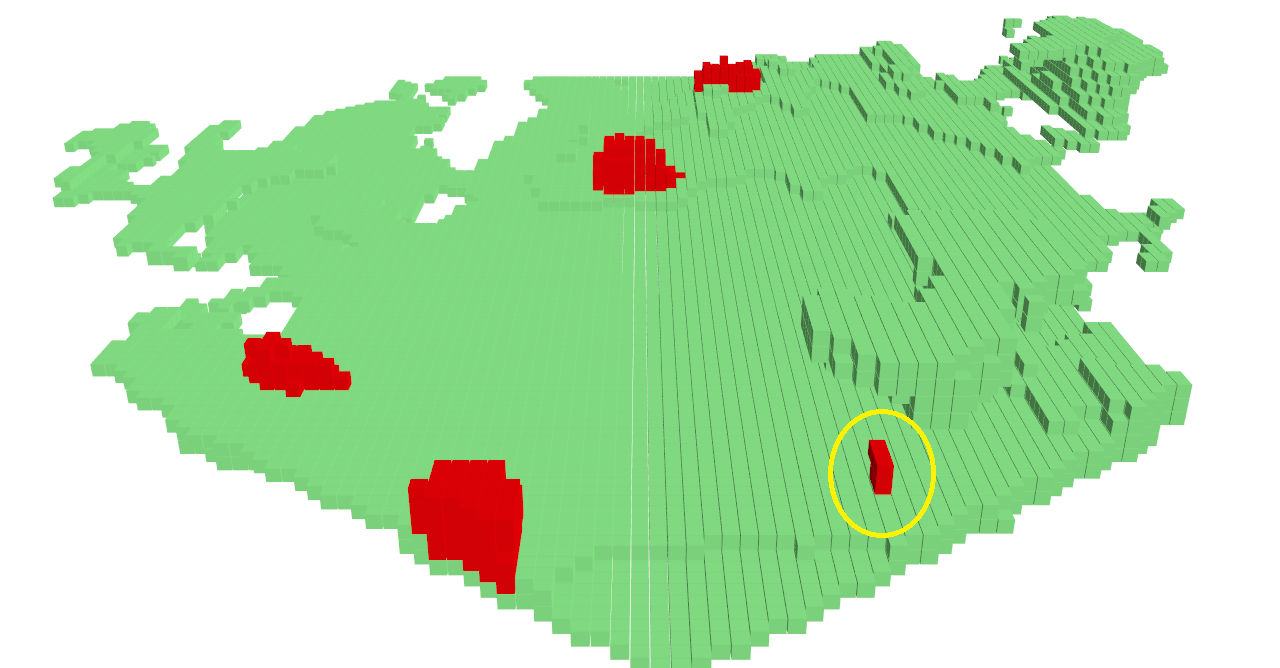}
        \caption{RadarOcc prediction}
        \label{fig:image3}
    \end{subfigure}
    \caption{Example of {\sysname} outperforming 32-line LiDAR on objects with low radar cross-section: the pedestrain is recognized.}
    \label{fig:small_object}
\end{figure}

\subsection{Example of failure cases}
We observed some failure cases of {\sysname} due to some reasons, such as insufficient resolution and decreased Signal-to-Noise Ratio (SNR) at far distances. An example of such failure cases is exhibited in Fig.~\ref{fig:falure_1}. We hope this could shed the light on future research in this field.

\subsection{How we handle object with low radar cross-section}
In our method, we address objects with low radar cross-section (RCS) from two key perspectives:

\textbf{Input perspective.} We utilize 4D radar tensor (4DRT) data instead of radar point clouds for 3D occupancy prediction. This approach avoids the loss of weak signal returns that can occur during the point cloud generation process, \emph{e.g.}, those filtered out by the CFAR detection, preserving more measurements from low RCS objects compared to radar point clouds.

\textbf{Method perspective.} Our sidelobe-aware spatial sparsifying technique selects the top-elements for each individual range rather than the entire dense RT. As shown in Fig.~\ref{fig:sparsifying}, this method retains critical measurements scattered across different ranges, including both low and high RCS objects. This contrasts with percentile-based methods, which often concentrate on elements corresponding to high RCS objects, thereby missing important data from low RCS objects. 

As a result, our method is effective in recognizing objects with low RCS, such as pedestrians, when predicting 3D occupancy. Figure~\ref{fig:small_object} shows an example where {\sysname} successfully handles low-RCS objects while 32-line LiDAR not.

\end{document}